\documentclass[12pt]{article}

\usepackage{amssymb,amsmath,amsfonts,eurosym,geometry,ulem,graphicx,caption,color,setspace,sectsty,comment,footmisc,caption,pdflscape,subfigure,array,enumitem}
\usepackage[draft,colorlinks=true,linkcolor=blue,citecolor=blue]{hyperref}

\usepackage{rotating}
\usepackage{mdframed}

\usepackage[utf8]{inputenc}

\newcolumntype{L}[1]{>{\raggedright\let\newline\\arraybackslash\hspace{0pt}}m{#1}}
\newcolumntype{C}[1]{>{\centering\let\newline\\arraybackslash\hspace{0pt}}m{#1}}
\newcolumntype{R}[1]{>{\raggedleft\let\newline\\arraybackslash\hspace{0pt}}m{#1}}

\newcommand*\pr[1]{{\mathbb{P}} \! \left(#1 \right)}

\newcommand*\dd{\text{d\kern 0.03em}}
\newcommand*\dz{\text{d\kern 0.05em}z_t}
\newcommand*\dq{\text{d\kern 0.05em}z_t^{\mathbb{Q}}}

\usepackage[
    backend=biber,
    maxcitenames=2,
    style=authoryear,
    natbib=true,
    url=true, 
    uniquename=false,
    doi=true,
    eprint=false
]{biblatex}
\usepackage{tabularx}
\usepackage{caption}
\newcommand{\re}[1]{\textcolor{blue}{[#1]}} 

\newcommand{\hi}[1]{} 

\newcommand{\colmarker}[1]{\multicolumn{1}{c}{#1}}

\usepackage{booktabs}

\usepackage[super]{nth}

\usepackage{titlesec}
\titleformat{\section}{\normalfont\large\centering}{\bfseries {\bfseries\thesection.}}{0.5em}{\MakeUppercase}
\titlespacing*{\section}{0pt}{2em}{0.25em}

\titleformat{\subsection}{\normalfont\bfseries\large\centering}{\thesubsection.}{0.5em}{}
\titlespacing*{\subsection}{0pt}{1em}{0.3em}{}

\titleformat{\subsubsection}{\normalfont\large\centering}{\thesubsubsection.}{0.5em}{}
\titlespacing*{\subsubsection}{0pt}{1em}{0.3em}{}

\begin{document}

\begin{titlepage}
\title{Applications of Machine Learning in Document Digitisation\thanks{Acknowledgements: We thank Peter Sandholdt Jensen, Joseph Price, and Michael Rosholm for useful comments. We also thank Søren Poder for contributing his expertise on digitisation of historical documents. We gratefully acknowledge support from Rigsarkivet (Danish National Archive) and Aarhus Stadsarkiv (Aarhus City Archive) who have supplied large amounts of scanned source material. We also gratefully acknowledge support from  DFF who has funded the research project ``Inside the black box of welfare state expansion: Early-life health policies, parental investments and socio-economic and health trajectories" (grant 8106-00003B) with PI Miriam Wüst.}}
\author{Christian M. Dahl\footnote{Department of Business and Economics, University of Southern Denmark}\footnotemark[2] $\quad$ Torben S. D. Johansen\footnotemark[2] $\quad$ Emil N. Sørensen\footnote{School of Economics, University of Bristol } \\ $\quad$ Christian E. Westermann\footnotemark[2] $\quad$  
Simon F.  Wittrock\footnotemark[2]}

\date{\today}
\maketitle
\vspace{-2em}\begin{abstract}
\noindent Data acquisition forms the primary step in all empirical research. The availability of data directly impacts the quality and extent of conclusions and insights. In particular, larger and more detailed datasets provide convincing answers even to complex research questions. The main problem is that “large and detailed” usually implies “costly and difficult”, especially when the data medium is paper and books. Human operators and manual transcription have been the traditional approach for collecting historical data. We instead advocate the use of modern machine learning techniques to automate the digitisation process. We give an overview of the potential for applying machine digitisation for data collection through two illustrative applications. The first demonstrates that unsupervised layout classification applied to raw scans of nurse journals can be used to construct a treatment indicator. Moreover, it allows an assessment of assignment compliance. The second application uses attention-based neural networks for handwritten text recognition in order to transcribe age and birth and death dates from a large collection of Danish death certificates. We describe each step in the digitisation pipeline and provide implementation insights. 

\bigskip
\end{abstract}
\setcounter{page}{0}
\thispagestyle{empty}
\end{titlepage}
\pagebreak \newpage


\doublespacing

\section{Introduction}
Big data have brought new opportunities in economic research \citep{varian2014tricks,einav2014economics}. However, big data are not confined to be contemporary. A recent review by \citet{gutmann2018bigdata} highlights that large collections of scanned historical documents are essential examples of big data, and \citet{gutmann2018bigdata} describe some of the challenges of harnessing such information in research. In particular, \citet{gutmann2018bigdata} mention the prospects of automated record linking by applying machine learning (ML) to transcribed records. However, they do not comment on the important opportunities of using ML to automate the data collection from the raw images.

Traditionally, historical data have been collected manually either by research assistants, using (possibly paid) crowdsourcing, or by complete outsourcing to a transcription company.\footnote{Amazon Mechanical Turk is an example of paid crowdsourcing where  workers are paid an amount every time they solve some pre-specified task. Alternatives such as Zooniverse provides the infrastructure to do crowdsourcing but otherwise rely on volunteers.} Manual data collection has limited scalability and this reduces the value of large scanned document collections as they are difficult to operationalise for statistical analysis. ML methods provide a potential solution to this problem. These methods are easily scalable to millions of documents, they are fast relative to human transcription, and provide reproducible results.

The economic literature does not have well-described applications of deep- and machine learning to collection of data from scanned documents. Similarly, while abundant with methods and models, the ML literature also lacks discussion of complete solutions to this problem, see e.g. \citet{nagy2016docs} for a general overview of document digitisation. Often the focus is on improving and benchmarking specific models in isolation using standardised datasets \citep[cf.][]{graves2008novel,bluche2014htr,lee2016recursive}. Such work is of limited practical use when implementing a complete data collection pipeline where the documents are non-standard and multiple models (e.g. for transcription and layout classification) need to work in unison.

The application and development of ML methods are not a foreign enterprise in economics. We aim to illustrate that ML-based data collection relies on methodology that should be familiar to contemporary economists and social science researchers \citep{varian2014tricks}. In recent years, there is a growing literature in econometrics on the applications of ML for estimating heterogeneous treatment effects \citep{wager2018estimation}, optimal policies \citep{athey2017efficient}, high-dimensional controls \citep{chern2018debiased}, instrumental variable selection \citep{windmeijer2019use}, and simulations \citep{athey2019using}. In addition, there is a strand of literature examining ML approaches to record linking \citet[cf.][]{abramitzky2019automated,bailey2017well,ruggles2018linking}. We identify that the capabilities of ML for collecting data from images is an important but missing component in this literature.

In this work, we contribute by describing two applications where we apply ML to collect data from scanned documents. Also, we discuss useful methodology implemented with freely available tools such as Python, PyTorch, Tensorflow, and OpenCV. This not only demonstrates the capabilities of ML in this context but also provides evidence that it is a worthwhile endeavour. We focus on scanned images with information in tables or forms as such documents can easily be mapped to the row-column data structures that are used in statistical analysis. We do not attempt to give an in-depth and exhaustive review of the existing ML methods -- the variety is simply too large for this to be meaningful. Also, we are not concerned with achieving state-of-the-art accuracy rates or undertaking systematic model comparisons. Rather, we focus on exemplifying how ML models can be adapted to solve actual data collection problems. In this respect, our results serve only as guidance and the performance could be substantially improved by proper hyperparameter optimisation, model tuning, and ensembles. 

The first application demonstrates that unsupervised layout classification can be applied to nurse journals to construct a treatment indicator. In the second application, we apply attention-based handwritten text recognition to transcribe age and birth and death dates from a large collection of Danish death certificates. We describe all steps in the transcription pipeline and suggest methods for layout classification and table segmentation. Finally, we also elaborate on the performance of ML compared to online crowdsourcing. 

The paper is organised into three main sections. Section~\ref{sec:nursesapplication} investigates how to construct the nurse treatment indicator. Section~\ref{sec:dcapplication} exemplifies the steps necessary to extract and transcribe data from the death certificates. Section~\ref{sec:conclusion} concludes.

\section{Early-life care in Denmark - Layout Classification}\label{sec:nursesapplication}
\begin{figure}\centering
\includegraphics[width=\textwidth]{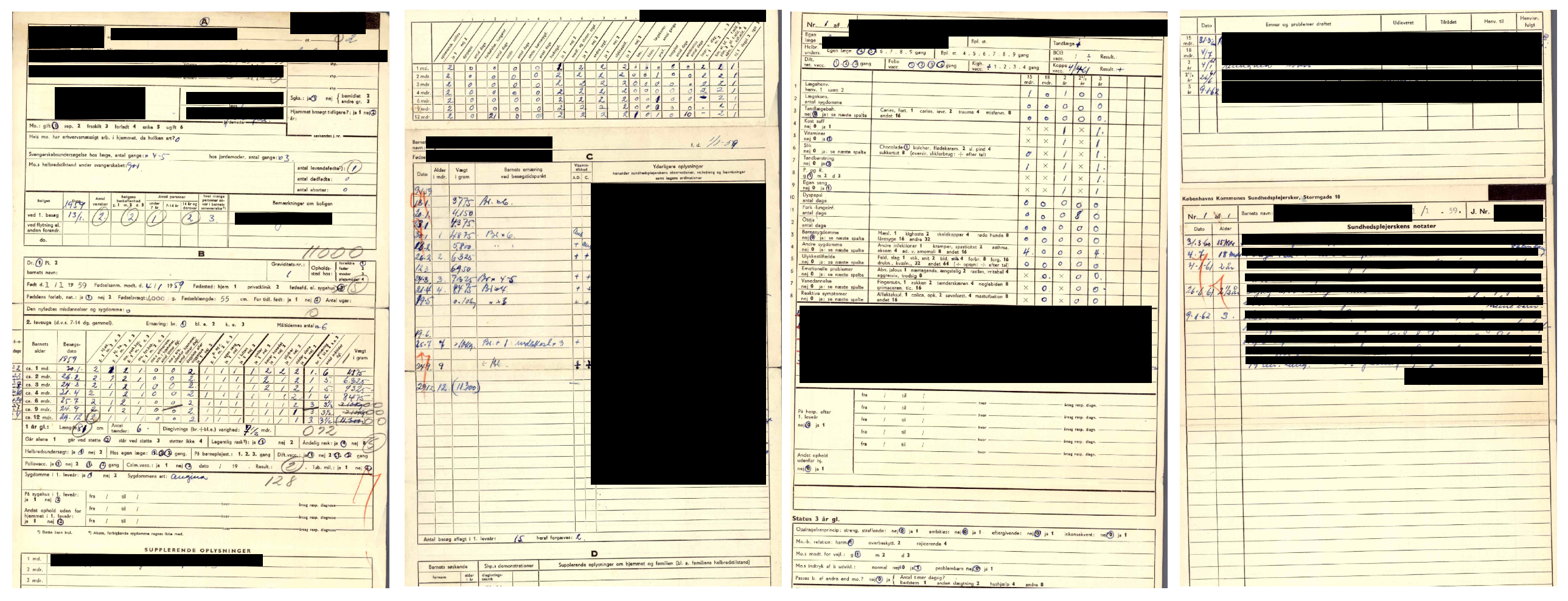}
\caption{Example of a typical nurse journal. The third page shows the treatment table. We are blacking out sensitive information.}
\label{fig:nursejournal}
\end{figure}

Interventions and estimation of treatment effects are central topics in both theoretical and applied economic research. However, prior to estimation, we need an assignment of each individual to a treatment or control group. Often treatment assignment is inferred based on an intervention or policy that (quasi) randomly has assigned each individual, e.g. \citet{krueger1991birthyear}. This section considers a policy where a subset of infants was made eligible to participate in an expanded care programme. The participants in the programme received additional home visits from nurses. Enrolment in the programme was governed by the date of birth. Individuals born in the first three days of each month were eligible to receive additional monitoring. The details of approximately $95,000$ infants (whether enrolled or not) were collected in journals kept by the health care system. The journals have previously been described and used by \citet{biering1980breast}, and \citet{bjerregaard2014effects} have manually transcribed a small subset of the contents to study birth weight and breastfeeding. The infants who received additional monitoring have a specific follow-up table in their journal only if the monitoring took place, i.e. the presence of the table is decided by actual treatment, not eligibility. The journals have been scanned and are available as digital images.\footnote{The journals have been made available through the DFF funded research project "Inside the black box of welfare state expansion: Early-life health policies, parental investments, and socio-economic and health trajectories" (grant 8106-00003B) with Miriam Wüst as PI.} While parts of the journals have previously been digitised, the presence of the treatment table was not recorded. Figure~\ref{fig:nursejournal} illustrates the pages in a typical journal. 

In the following, we construct a treatment indicator using unsupervised ML by analysing the layout of each journal page and thereby identifying the group of children that actually received follow-up care. We compare this ML-based detection to an intention-to-treat (ITT) indicator inferred from the three-day policy and find that there is non-compliance. This illustrates that statistical models applied to images can, even without transcription, provide important information in applied economic research.

Our dataset contains $95,313$ journals with a total page count of $261,926$. The number of pages per journal is random, and the ordering of the pages is not fixed. This implies that all pages in all journals have to be reviewed to identify the treated.

\subsection{ML detection of treated individuals}
Since the treated individuals can be identified by the presence of a particular page in their journal, we can use layout classification to detect treatment. If a page in their journal is classified as having the treatment table layout, then the individual is classified as treated. We did not have access to a labelled dataset to train a supervised classifier for the treatment page -- as will often be the case in practice. Thus, we pursue an unsupervised approach where we rely only on the scanned images without labels. Note that we still need to manually construct an evaluation dataset to probe the performance of the applied method. However, if the unsupervised method is found to perform sub-par, then the evaluation set can serve as the basis for training a supervised classifier. In this sense, any manual transcription is not wasted. We will show an example of a supervised classifier for the same purposes in Section~\ref{sec:dcapplication}.

The documents are scanned and stored digitally as image files. Images consist of coloured dots called pixels. Each pixel is characterised by a location and a colour. Stacking a certain number of pixels horizontally and vertically forms an image. Thus, we can consider an image of $h \times w$ pixels to be a $h \times w$ matrix where each entry corresponds to a single pixel. In grayscale photos, each pixel can only attain white, black, and shades in-between. This is represented by a byte (8 bits) specifying a value between 0 and 255 with 0 being black and 255 white. The core of the machine digitisation process can be formulated as various statistical learning problems where we model different aspects of the visual information to learn a mapping from the image matrix into a representation that is suitable for economic analysis. Learning this mapping represents an array of challenges. In particular, this is often challenging because the image matrix can be of very high dimension. A feature is a lower-dimensional variable that captures some aspect of the high-dimensional image, and hopefully in a way that is more informative than the raw image data itself. Convolutional neural networks, see \citet[Chp. 9]{goodfellow2016deep}, learn to extract such features when trained for image classification and it turns out that these features are generally informative despite being trained on a specific dataset \citep{simonyan2015vgg16}. This property is exploited in transfer learning where parts of a neural network trained on one set of images are applied for a new task on a different set of images \citep{pan2009transfer}. The VGG16 network is an example of a deep convolutional neural network that was trained on over $1$ million photos to distinguish between $1,000$ objects \citep{simonyan2015vgg16}. Based on the concept of transfer learning, we use this pre-trained network to extract features from the journal images. This is a useful trick that can provide informative features without the need to train more sophisticated feature extractors or models. It works similar to traditional feature extractors, e.g. SIFT \citep{lowe2004distinctive} and SURF \citep{bay2008speeded}, but the feature representation is learned instead of manually engineered. The classification part of VGG16 is discarded -- we do not care about the original classification task -- and we only keep the convolutional network. Each journal page is passed through the VGG16 convolutional network and we obtain a $512$-dimensional feature vector that describes some aspects of the visual information.

\begin{figure}\centering
\includegraphics[width=0.65\textwidth]{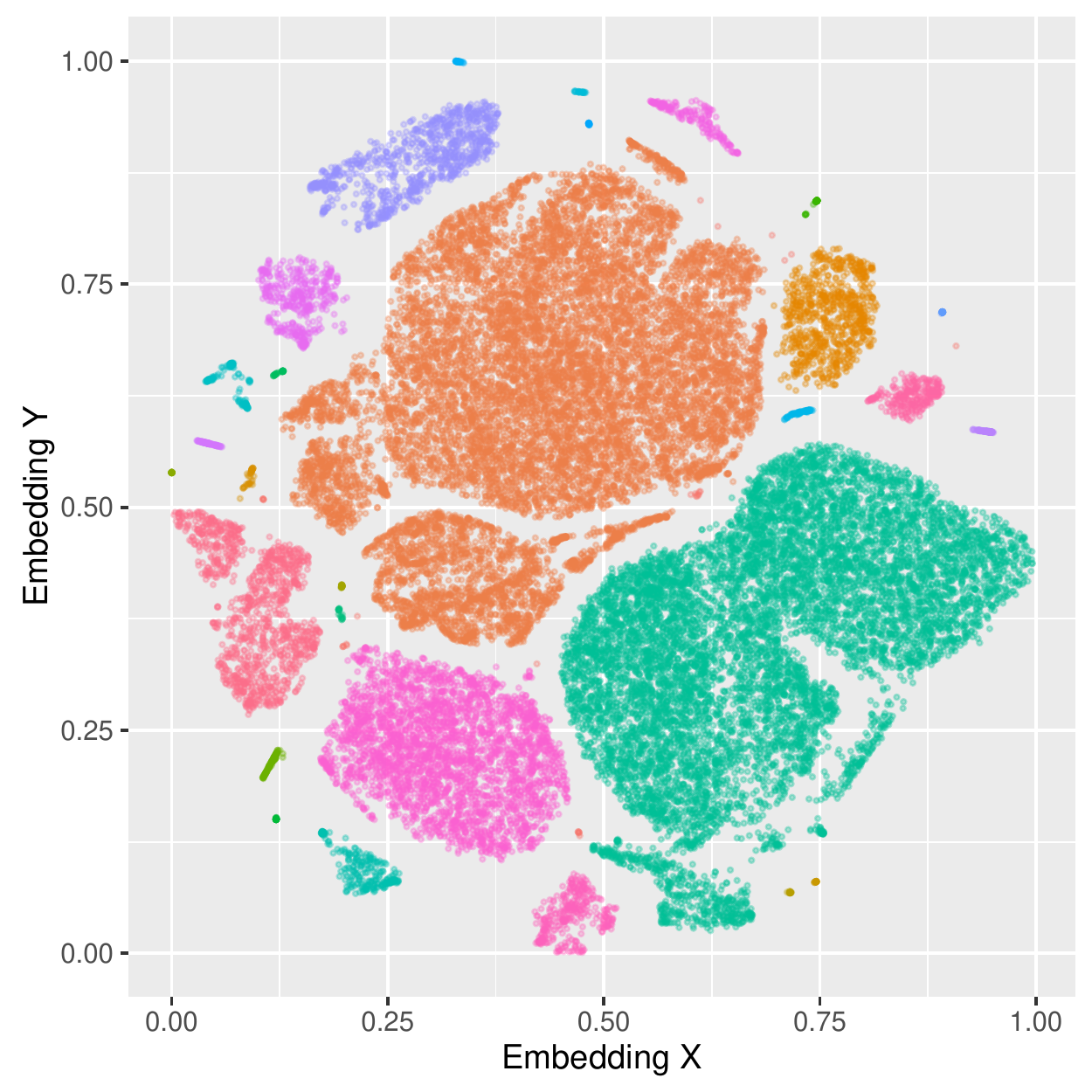}
\caption{$2$D t-SNE visualisation of the feature space of the journal pages. Each point represents a journal page and the colours correspond to the labels assigned by the clustering algorithm. Pages with similar layout cluster together. The embeddings have been subsampled to reduce cluttering, so only $30,000$ randomly sampled embeddings are displayed. There is a total of $37$ clusters which are manually annotated. The treatment pages are contained in four clusters.}
\label{fig:nurse_embeddings}
\end{figure}

Next, we use unsupervised methods to explore the features. The features are clustered using DBSCAN \citep{ester1996dbscan} -- a density-based clustering algorithm. Pages with similar layout should cluster together as they share a similar VGG16 feature vector. To visualise the features and clusters, we embed them in a two-dimensional space with t-distributed Stochastic Neighbour Embedding (t-SNE) \citep{maaten2008tsne}. The t-SNE method provides a convenient way to visualise high-dimensional spaces in two or three dimensions while retaining the local structure between points, i.e. points that are close in the high-dimensional space also tend to be close in the low-dimensional embedding space. The t-SNE embeddings for the nurse records are depicted in Figure~\ref{fig:nurse_embeddings}. Each point represents a page in a journal and the points are colored according to their assigned cluster. A clear structure is evident. There are 37 clusters which we manually review and annotate according to their contents. Annotation is carried out by randomly sampling 10 pages from each cluster and assigning a label for the whole cluster based on the contents of these pages. This amounts to 370 journal pages that need manual review (out of 261,926). This procedure shows that the treatment pages are contained in four distinct clusters. We extract all pages residing in these clusters and classify the underlying individuals as treated, i.e. an individual is treated if any page in their journal belongs to one of the four treatment page clusters.

\begin{table}\centering
\begin{tabular}{l rrrr }
\toprule\toprule
& \multicolumn{2}{c}{ML detection} \\ \cline{2-3}
Ground truth & Treated & Not Treated \\ \midrule
Treated & 234 & 0 & 234 \\
Not Treated & 0 & 3766 & 3766 \\
 & 234 & 3766 \\
\bottomrule\bottomrule
\end{tabular}
\caption{Confusion matrix for the ML treatment detection model. The frequencies are based on a randomly sampled and manually reviewed validation set of $4,000$ journals ($10,914$ pages). The treatment detection model does not rely on any segmentation, but detects the presence of the whole page containing the treatment table.}
\label{table:confusionnurses}
\end{table}

\begin{table}\centering
\begin{tabular}{l rr }
\toprule\toprule
& Policy detection (ITT) & ML detection \\ \midrule
Treated & 7,912 & 5,735 \\
- Born 1st-3rd & 7,912 & 4,247 \\
- Born 4th-31st & 0 & 455 \\[0.5em]
Non-compliers & - & 4,120 \\
\bottomrule\bottomrule
\end{tabular}
\caption{Treatment indicator. Policy assignment is based on an official assignment rule which offered all children born in the first three days of each month to enroll in the nurse visiting programme. The ML assignment is based on the machine learning model and bases assignment on the presence of the treatment page in the journals. This allows for assessment of compliance in addition to the intention-to-treat effect, i.e., the date-of-birth assignment mechanism.}
\label{table:treatmentnurses}
\end{table}

To evaluate the unsupervised procedure, we manually constructed a ground truth evaluation dataset by reviewing all $10,914$ pages of $4,000$ randomly selected journals. For each journal, we recorded the presence of the treatment table. The dataset was reviewed twice and $234$ journals with treatment were found.
Table~\ref{table:confusionnurses} provides a confusion matrix for the ML treatment detection in the evaluation sample. All 234 treated and 3,766 untreated individuals are correctly classified by DBSCAN with zero false positives/negatives despite heavy class imbalance. Note that the classifier could obtain an accuracy of $3,766/4,000 \approx 94.15\%$ by simply assigning everyone as non-treated.\footnote{Actually, the class imbalance is even more severe. The ML model classifies pages and in the evaluation set only $234/10,914 \approx 2.14\%$ pages contain the treatment table. In light of this, the high recall of the model is especially satisfying.} Obviously, this is not desirable and highlights the need for other performance measures. An option is to consider the precision and recall \citep[p. 184--185]{murphy2012machine}. In our context, precision is \emph{the number of individuals predicted as treated that are truly treated}, while recall is \emph{the number of individuals detected as treated compared to the total number of treated}. These two measures are especially relevant in detection and retrieval tasks such as those considered here \citep[cf.][p. 185]{murphy2012machine}. In the results from the unsupervised method, both precision and recall are unity. While this is highly satisfactory, it is only an estimate of the true performance as we only use a subset of the data for evaluation. It is conceivable that the method can make some mistakes across the whole collection of $261,926$ pages.

Apart from the performance of the classifier itself, the results from the layout detection provide valuable insights on treatment assignment. From Table~\ref{table:treatmentnurses} it is evident that not all eligible individuals received the follow-up visits. 7,912 children are eligible but only 4,247 individuals born in the three-day eligibility period actual received a visit.\footnote{Although $7,912$ individuals born between the 1st and the 3rd appears low when considering the sample size of roughly $95,000$ children, we observe birth date for only $84,659$ children. Hence,  $9.35\%$ of these children are born between the 1st and the 3rd, which is still slightly lower than expected ($9.86\%$).} This is a treatment uptake of only $53.68\%$. In addition, there is a group of $455$ children born outside the intervention days that still receive treatment even though the policy assigns them as controls.\footnote{A search for more information on these cases is ongoing.} This reveals an issue with non-compliance ($3,665 + 455 = 4,120$ non-compliers) which might have implications when estimating treatment effects, see e.g. \citet{angrist1996identification}. The performance of the ML detection is very encouraging, and in addition the ML approach also reveals details about the intervention that would otherwise have been lost, unless the whole collection of $261,926$ pages was manually reviewed.

Keep in mind that the nurse journals are very uniformly scanned. Documents with more variation in quality might benefit from a supervised approach. For example, we found that the unsupervised method did not generalise well to the death certificates (Section~\ref{sec:dcapplication}). 

\begin{sidewaysfigure}[p]\centering
\includegraphics[width=1\textwidth]{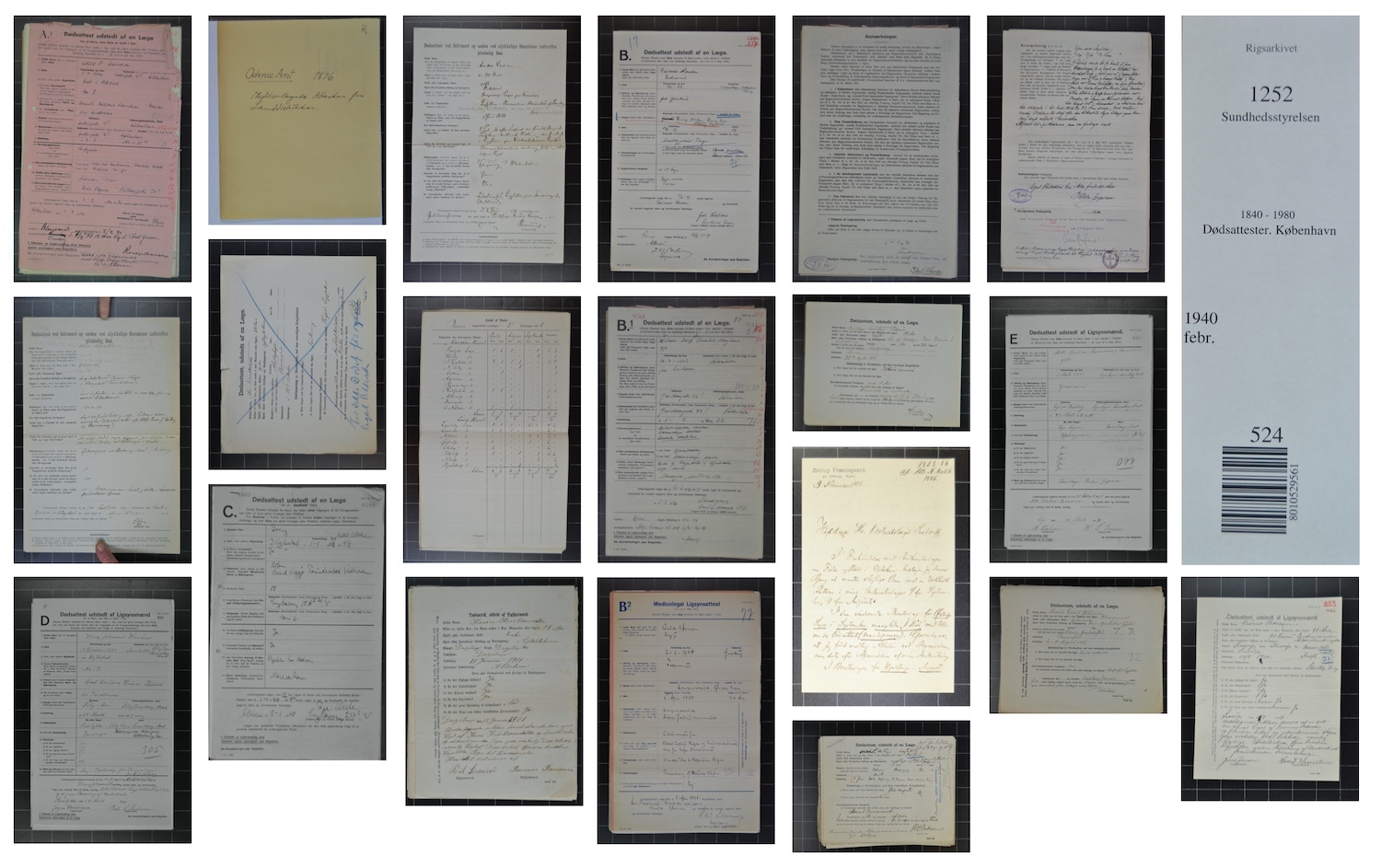}
\caption{A sample of the different pages in the collection of death certificates. The examples are not exhaustive.}
\label{fig:certificatetypes}
\end{sidewaysfigure}

\begin{figure}\centering
\includegraphics[width=0.8\textwidth]{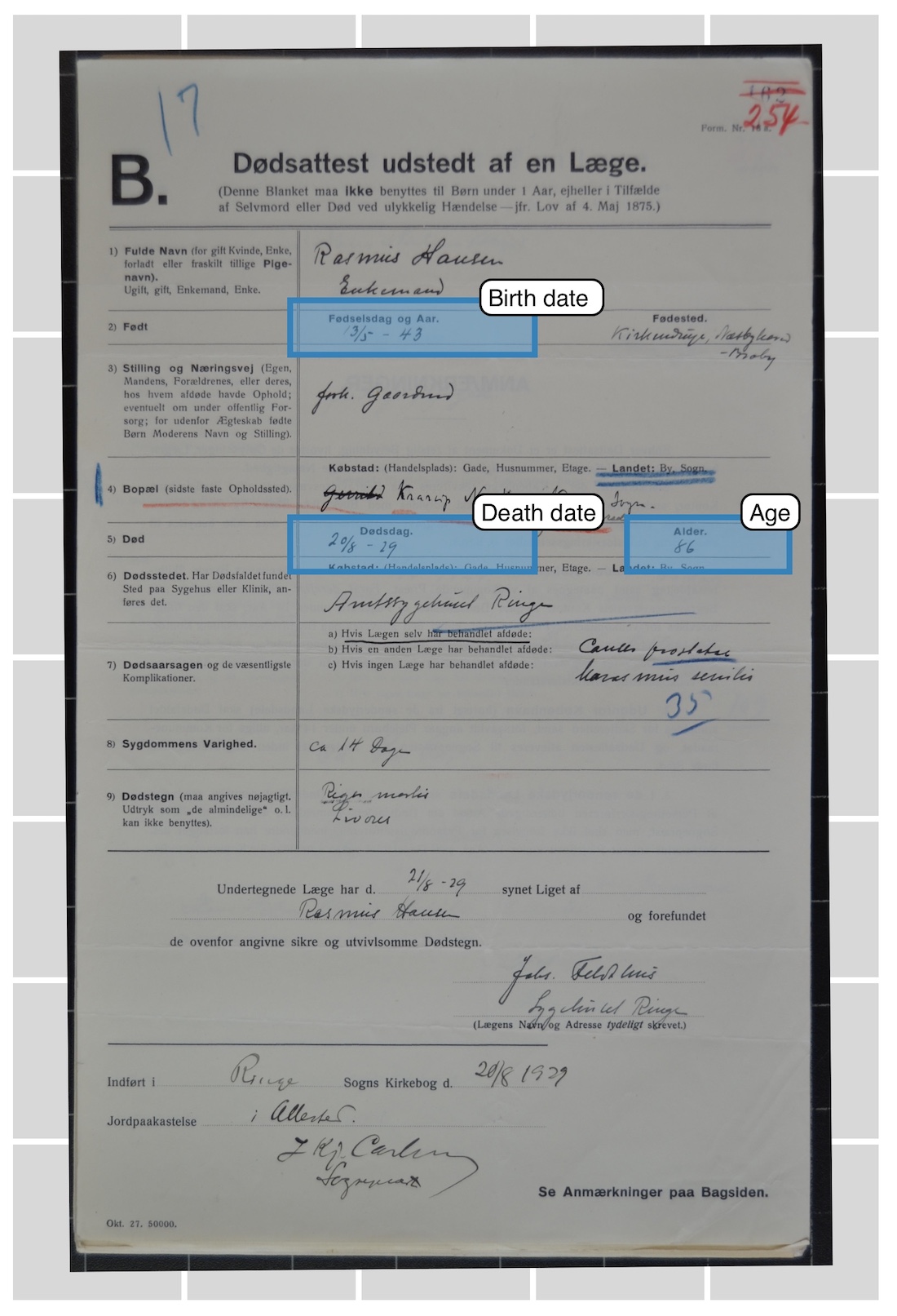}
\caption{The birth date, death date and age fields on a type B death certificate.}
\label{fig:typebfields}
\end{figure}

\begin{sidewaysfigure}[p]\centering
\includegraphics[width=1\textwidth]{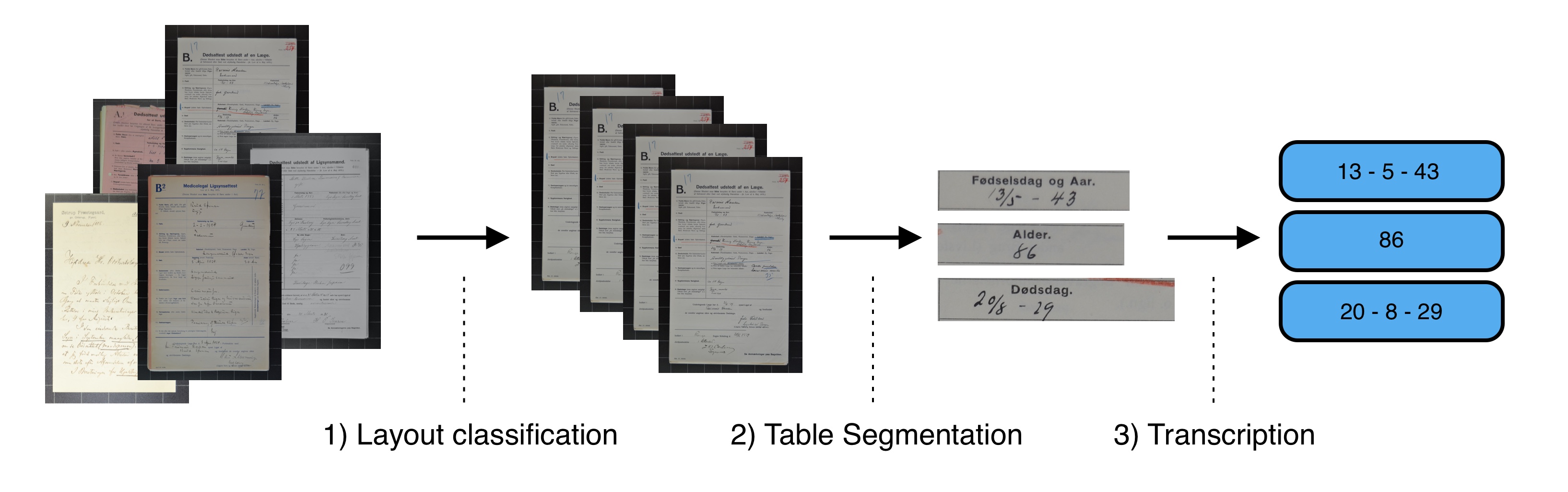}
\caption{The three steps in the ML data collection pipeline. (1) the collection of source images is sorted by layout. (2) the form in the document of interest is segmented into field images. (3) each field image is transcribed into a digital string.}
\label{fig:pipeline}
\end{sidewaysfigure}

\section{Mortality in Denmark - Handwritten Text Recognition}\label{sec:dcapplication}
In Denmark, the use of death certificates was introduced at the national level in 1832.\footnote{See the description at the Danish National Archive (in Danish) \url{https://www.sa.dk/da/hjaelp-og-vejledning/rigsarkivets-online-vejledninger/doedsattester-kom-godt-i-gang/}} A death certificate documents the death of a single individual and contains a table with fields for name, birth date, cause-of-death etc. The documents are stored on paper at the Danish archives, one certificate is one page. Due to privacy restrictions, the publicly available death certificates are restricted to the timespan 1832--1942. The Danish National Archive and volunteers have scanned a large collection of certificates and made them available online as digital images. Around 2.5 million death certificates are available for download and with approximately 10--12 fields per certificate; this amounts to 25--30 million individual fields to transcribe. Also, additional death certificates are continuously being scanned and added to the collection. We have a subcollection of approximately $250,000$ death certificates across multiple years and locations. These are not randomly sampled but reflect the order in which the archives scanned the documents. This is not a major drawback as our main purpose is to demonstrate the ML methods.
The certificates are based on pre-printed forms where the respective fields are filled in by hand. Due to the pre-printed form, they are characterised by a fairly consistent layout. However, the form changes structure over time and several subtypes of forms are used to distinguish between deceased infants, suicides, accidents, etc. These numerous subtypes are illustrated in Figure~\ref{fig:certificatetypes}. The de-centralised scanning of the documents -- multiple archives with different volunteers and equipment -- also produces significant variation in scan quality. The available fields on each certificate are -- with some variation -- name, marital status, birth date, death date, age, occupation, birth location, death location, cause of death, duration of disease, and clinical signs of death. 
In our application, we focus on transcription of dates (birth and death) and age. These fields have a well-defined dictionary and provide internal consistency, i.e. birth and death dates have to match with age. We limit our attention to the type B death certificate for two reasons: (1) the pre-printed form has clearly delineated fields and (2) our collection has a significant proportion of this type. Our layout model -- to be discussed later -- predicts that $44,903$ out of $250,000$ certificates are type B. Figure~\ref{fig:typebfields} shows a type B certificate with manual highlighting of the three relevant fields. 

We split the transcription process into three steps: 

\begin{enumerate}[label=(\arabic*)]
\item \textbf{Layout classification}: Separate the type B death certificates from the other types.
\item \textbf{Table segmentation}: Extract an image for each of the selected fields in the pre-printed form.
\item \textbf{Transcription}: Transcribe the extracted field images for birth date, death date, and age.
\end{enumerate}

\noindent The sequence of steps (1)-(3) is called the ML pipeline, see the illustration in Figure~\ref{fig:pipeline}. The pipeline consumes raw images of death certificates and produces string transcriptions for each field without human interaction. In the following sections, we describe each of the pipeline steps and evaluate the performance. Section~\ref{sec:dc:datasets} describes the transcribed datasets used for training and evaluation. The pipeline steps are considered in Sections~\ref{sec:dc:layout}--\ref{sec:dc:transcription} while Section~\ref{sec:dc:crowd} compares the ML pipeline to crowdsourced transcription.

\subsection{Overview of datasets}\label{sec:dc:datasets}
We train and evaluate three separate models for the ML pipeline. One model for layout classification and the two others for transcription. In this process, we rely on several datasets which are outlined in Table~\ref{table:datasets} and described in detail below. The table segmentation method does not need training so there is no dataset for this step.

\begin{enumerate}[label=(\arabic*)]
\item \textbf{Dates}: A training dataset with $11,630$ dates and an evaluation dataset containing $1,000$ dates. Data is approximately balanced between birth and death dates. The datasets are used to train and evaluate the transcription model for birth and death dates. Images are $320 \times 50$ pixels and the ground truth transcriptions are stored as strings in a standardised format. See Figure~\ref{fig:dateexampledataset}.

\begin{center}
\includegraphics[width=0.55\textwidth]{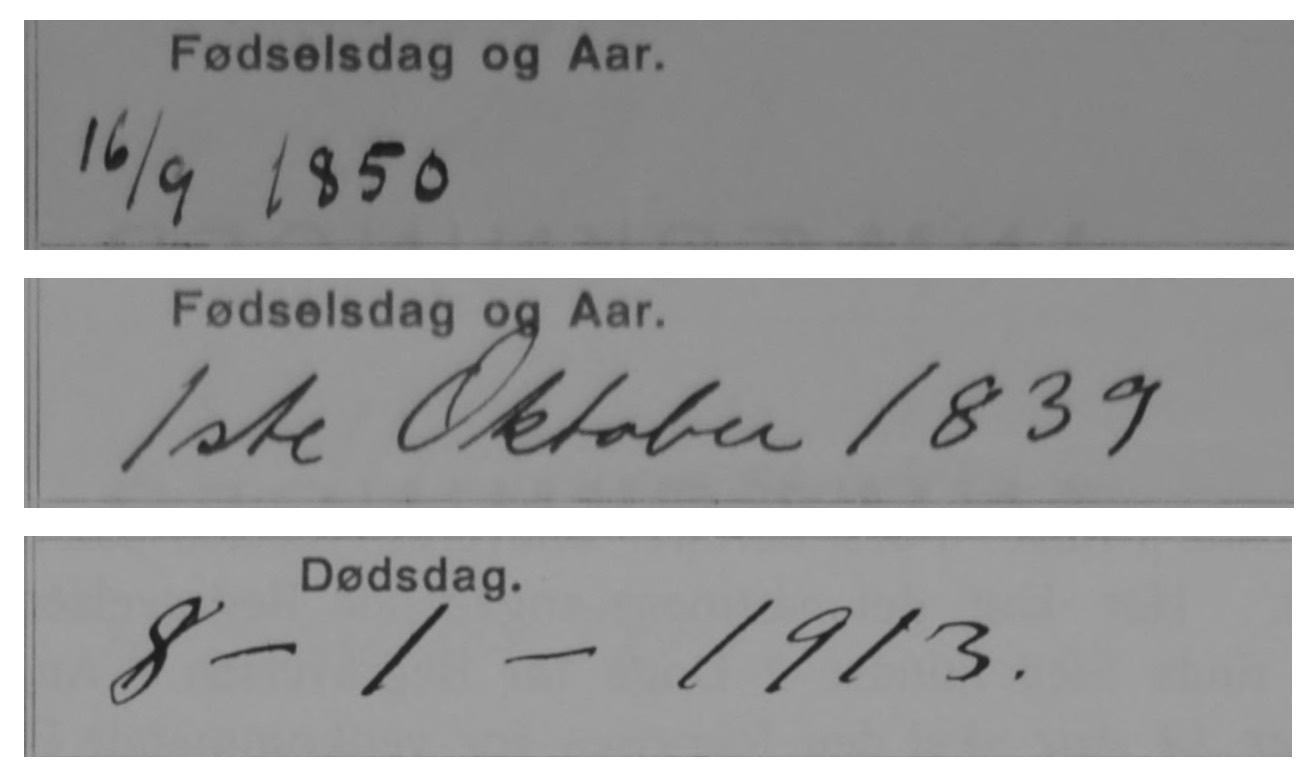}
\captionof{figure}{Examples of the images in the date datasets.}\label{fig:dateexampledataset}
\end{center}

\item \textbf{Ages}: The training and evaluation datasets contain $11,072$ and  $1,000$ ages respectively. The datasets are used to train and evaluate the transcription model for age. Images are $230 \times 75$ pixels and the ground truth transcriptions are stored as strings and exclude the age suffix, i.e., years, months, days, or hours. Non-integer ages are also excluded. See Figure~\ref{fig:ageexampledataset} for examples.

\begin{center}
\includegraphics[width=0.6\textwidth]{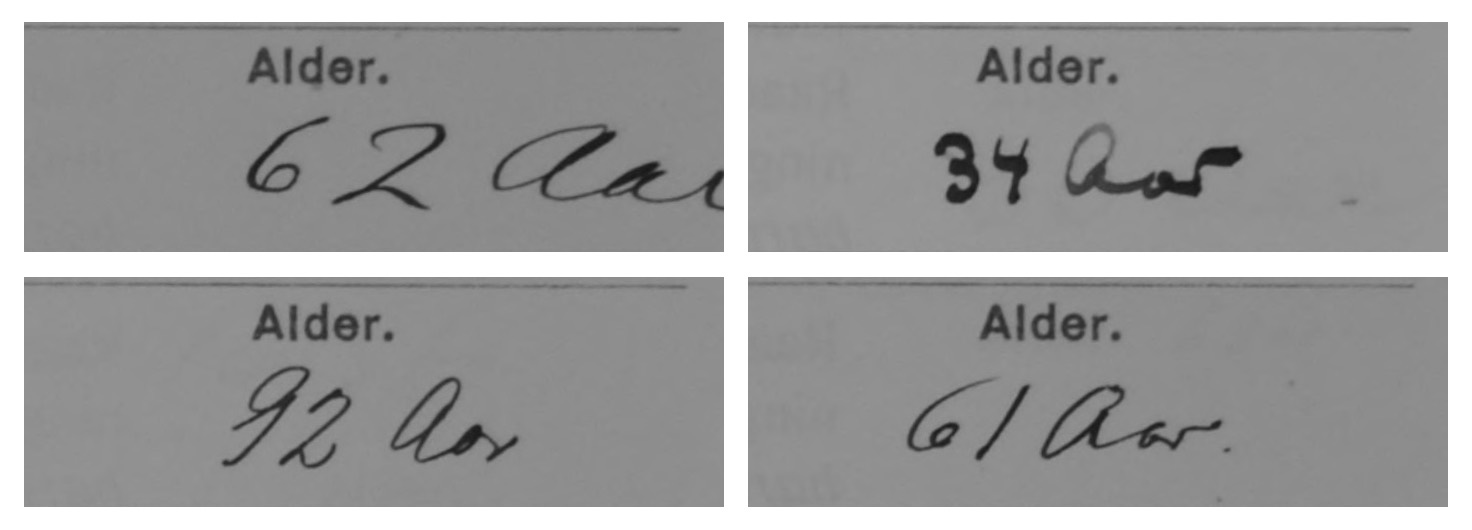}
\captionof{figure}{Examples of the images in the age datasets.}\label{fig:ageexampledataset}
\end{center}

\item \textbf{Layouts}: The training and evaluation datasets contain $7,000$ and $2,184$ pages respectively. They are used to train and evaluate the layout classification model for detecting certificates of type B. The images are of varying size and the ground truth layout type is stored as an indicator variable. The images are similar to those shown in Figure~\ref{fig:certificatetypes}.
\item \textbf{Crowdsourced dates}: A dataset containing 23,263 complete death certificates that intersect with our collection of death certificates. Transcriptions are only available for birth and death dates, not for age. 
This dataset is used for evaluating the end-to-end performance of the pipeline. 
\end{enumerate}

\begin{table}\centering
\begin{tabular}{l rr l}
\toprule\toprule
	& Training & Evaluation & Image size \\ \midrule
Dates & 11,630 & 1,000 & $320 \times 50$ pixels \\
Ages & 11,072 & 1,000 & $230 \times 75$ pixels \\
Layouts & 7,000 & 2,184 & Variable \\ 
Crowdsourced dates & - & 46,526 & $320 \times 50$ pixels \\
\bottomrule\bottomrule
\end{tabular}
\caption{Overview of the number of samples in each of the training and evaluation datasets used in the ML pipeline}
\label{table:datasets}
\end{table}

\noindent 
The evaluation and training datasets (1)-(3) are constructed by manually transcribing a random sample of field images from the death certificates. They have been verified twice by different individuals and images with segmentation errors have been removed. We ensure there is no overlap between the training and evaluation datasets.

The crowdsourced dataset (4) is freely available online from the Danish National Archive. Anybody can contribute to these transcriptions by editing them through an online interface.\footnote{See (in Danish) \url{https://bit.ly/2VnFLzb}}

\subsection{Step 1: Layout Classification}\label{sec:dc:layout}
\subsubsection{Method}

Layout classification refers to the process of organising a collection of documents by common layout structure, e.g. a common table, heading, or pre-printed landmarks. \citet{chen2007survey} provide a relevant but methodologically outdated introduction to this problem. The layout type is important as the table segmentation step relies on a pre-defined template that must match with the pre-printed structure in the image. Due to variations in layout across the collection of death certificates (see Figure~\ref{fig:certificatetypes}), we need to construct one template for each type. Every time we fit a template to a certificate, we need the template and certificate types to match. This is straightforward if the certificates are sorted according to layout, as in this case, all images in a given class will share the same template.

As we saw in Figure~\ref{fig:certificatetypes}, an image $\textbf{X}_i$ of a death certificate can belong to one of $K$ layout types that can be distinguished visually, e.g. ``Type B'', ``Type A'', and so on. Let the layout type of image $i$ be $Y_i = k$ with $k = 1, 2, ..., K$. 
The $K$ types do not need to mimic the visual types in the documents, we can easily focus on ``Type B'' and label everything else as ``Other''. We are interested in learning a model for the probability $\pr{Y_i = k \, | \, \mathbf{X}_i}$ such that we can infer the most likely layout type $\hat{y}_i = \arg \max_k \pr{Y_i = k \, | \, \mathbf{X}_i}$. 
This resembles a conventional $K$-class classification problem with the only difference that $\mathbf{X}_i$ is an image.

We already saw an example of unsupervised layout classification in the context of the nurse journals in Section~\ref{sec:nursesapplication} where we emphasised the importance of constructing a lower-dimensional feature that describes the raw image. In the supervised setting, there are two main considerations: (1) what features should we use, i.e. how do we construct a feature $g(\textbf{X}_i)$ that represents the high-dimensional information in $\mathbf{X}_i$, and (2) what classifier should we use, i.e. what model do we choose for the probability $\pr{Y_i = k \, | \, g(\mathbf{X}_i)}$. 
This cannot be answered definitively and depends on the application. There are many equally viable approaches. In the nurse journal application, we solved (1) by using the features from a pre-trained neural network (VGG16) and used these directly in a cluster analysis. A modern supervised end-to-end approach would be to train a convolutional neural network (CNN) to classify the pages using raw images as input, i.e. $g(\textbf{X}_i) = \textbf{X}_i$. In this case, the neural network will solve both (1) and (2) as it learns both the feature extractor and the classifier during training. 

We take a simpler approach and use the visual Bag-of-Words (BoW) method to classify the layout type of the certificates. This method provides an intuitive definition of features in terms of visual landmarks or ``words''. 
Bag-of-words is a technique originally developed to classify chunks of text according to the content \citep[p. 87]{murphy2012machine}. It operates by determining the frequency of words (i.e. the frequency of some global set of words -- the bag-of-words) in each text document. The method has been applied successfully in the field of computer vision, see \citet{csurka2004visual,sivic2009efficient}, where we instead operate with a bag of visual words, i.e. chunks of images.

The visual bag-of-words model creates features on the basis of a codebook or dictionary. This dictionary is constructed by extracting key points from a training dataset and clustering them such that we obtain $M$ groups of key points where key points within a group appear similar in some sense. If we think of a training dataset that consists of photographs of animals, we could have a key point cluster related to eyes, one for ears, etc. Each of these clusters is a \emph{visual word}. The visual words are similar to the feature clusters we discovered in the nurse journals using the unsupervised method, i.e. those in Figure~\ref{fig:nurse_embeddings}. Here we extract the features using Speeded-Up Robust Features (SURF) \citep{bay2008speeded} as this has historically been the common choice, see \citet{csurka2004visual}. 

When the dictionary has been constructed, we are ready to create the actual feature vectors for the images. For a given training image $i$, we extract SURF key points and assign each of these to the $M$ key point clusters based on distance. We then count the number of features from the image that belongs to each of the $M$ key point clusters and construct a vector of normalised frequencies which serves as the final feature vector of the image. Note that the size $M$ of the dictionary determines the dimensionality of the final feature vector.
In the classification step, we train a model to classify each feature vector (and thus each image) into one of the $K$ classes. The classifier can be of any type. Here we use support vector machines with radial kernels, see the introduction in \citet[Chp. 9]{james2013introduction}. There are several tuning parameters in the BoW model (dictionary size, margin etc.) which we selected using leave-one-out cross-validation. 
The computational burden lies in the initial keypoint extraction and clustering. When the appropriate features have been extracted, it is very fast to fit the actual classifiers and hence it is computationally fast to do cross-validation for the classifier hyperparameters\footnote{Note that for large training sets, runtime for support vector machines with radial basis kernels increases from $O(n^2)$ to $O(n^3)$ and another classifier should be used.}. For an end-to-end neural network it is substantially slower to do hyperparameter optimisation as the whole network will need to be re-trained -- possibly for hours -- for each set of hyperparameters. This highlights that simpler classifiers can be useful as a first step before developing more complex models.

\subsubsection{Results}
\begin{table}\centering
\begin{tabular}{l rrrrr}
\toprule\toprule
& \multicolumn{4}{c}{ML detection} \\ \cline{2-5}
Ground truth & Empty & Other & A & B \\ \midrule
Empty 	& 13 		& 1 		& 0 		& 0 		& 14 \\
Other 	& 2 		& 1,535 	& 0 		& 0 		& 1,537 \\
A	 	& 0 		& 0 		& 109 	& 0 		& 109 \\
B	 	& 0 		& 0 		& 0 		& 524 	& 524 \\
 & 15 & 1,536 & 109 & 524 \\
\bottomrule\bottomrule
\end{tabular}
\caption{Confusion matrix for the BoW layout classification. The frequencies are based on a randomly sampled and manually reviewed evaluation set of $2,184$ death certificates. Death certificates are classified into four classes: Empty, A, B and Other. In this application we are only interested in the type B certificate.}
\label{table:dclayoutconfusion}
\end{table}

The classifier is trained on the layout dataset which contains $7,000$ full page images of randomly sampled death certificates together with an indicator for their ground truth layout type. We consider $K = 4$ distinct layout classes: (1) type A certificates, (2) type B certificates (our focus), (3) all other certificates and (4) empty pages.\footnote{The other classes (i.e. type A and empty) were used for a different application we do not discuss here.} We evaluate the classifier on the corresponding layout evaluation dataset ($2,184$ certificates) and the confusion matrix is given in Figure~\ref{table:dclayoutconfusion}. From the confusion matrix we see that the classifier provides highly satisfactory performance with only two false positives and one false negative in document classes we are not interested in. For the type B certificates the class-wise precision and recall are unity. We use the trained BoW classifier to predict across the entire collection of $\sim 250,000$ death certificates and find that $44,903$ certificates are type B. These are extracted for the table segmentation step.

\subsection{Step 2: Table Segmentation}\label{sec:dc:segmentation}
\subsubsection{Method}
In the table segmentation step, the goal is to extract smaller images corresponding to each field (or cell) of a larger form (or table) in the source image. \citet{couasnon2014tables} provide a general introduction to the topic. There are two components to this problem: (1) identifying the structure of the table in the source image, and (2) exploiting the structure to extract the field images. We apply a simple template-based approach where a predefined template is fitted over the source image using a set of landmark points.

In ongoing work, we are considering how to solve (1) using edge-detection neural networks. However, in this paper we rely on standard filtering operations from the computer vision literature to binarise the source image and find straight horizontal and vertical lines. This implies that we are not using ML and since the operations are application dependent we only briefly discuss this part in the results section. We solve (2) using point set registration methods where we apply the Coherent Point Drift (CPD) method of \citet{myronenko2010point}. This method relies on a probabilistic model where we learn a transformation between two sets of points using maximum likelihood and use it to directly fit a template to the source image.

\begin{sidewaysfigure}[p]\centering
\includegraphics[width=0.9\textwidth]{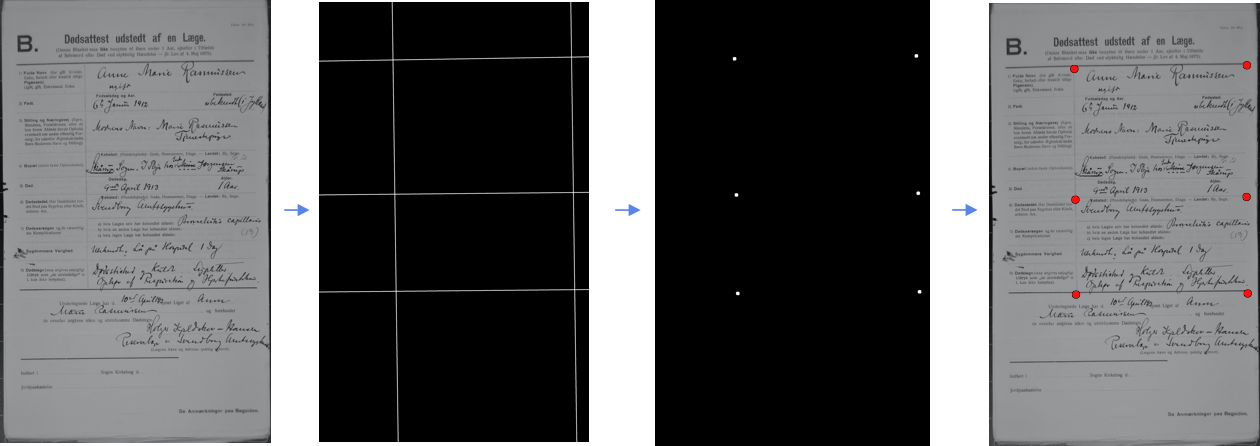}
\captionof{figure}{Landmark detection using morphological operations and a corner detector.}\vspace{1em}
\label{fig:dclandmarks}
\includegraphics[width=0.9\textwidth]{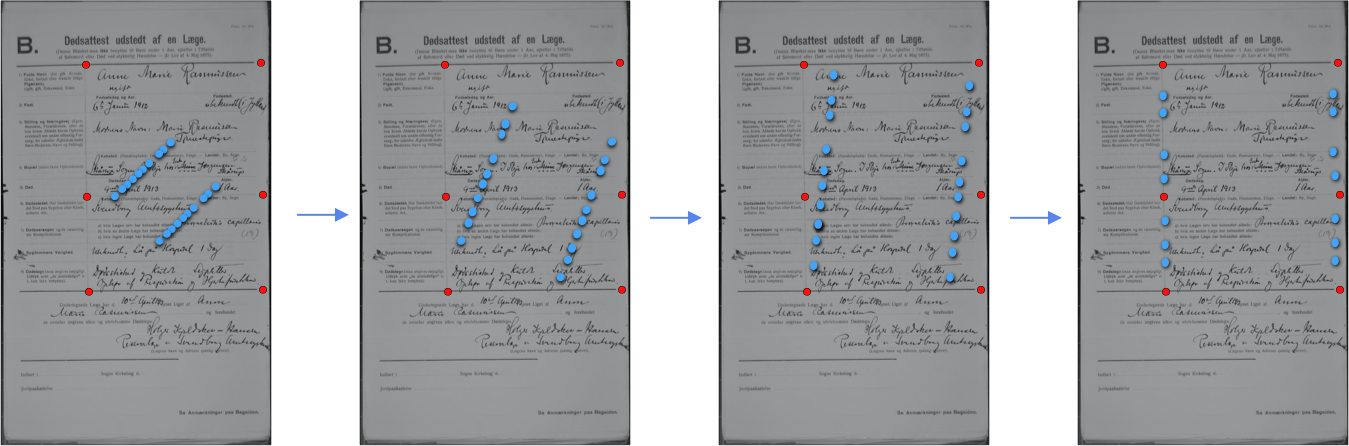}
\captionof{figure}{Point set alignment. Blue dots represent the template while red dots are the landmarks found in the source document.}
\label{fig:pointsetalignment}
\end{sidewaysfigure}

A point set is -- as the name implies -- simply a set of points. Since we are working with images, these points reside in the plane and are characterised by two coordinates. Let $P_1 = \{ (x_{11}, y_{11}), (x_{12}, y_{12}), ..., (x_{1N}, y_{1N}) \}$ be a set of points that define a template (e.g. the corners of a table) and analogously let $P_2 = \{ (x_{21}, y_{21}), (x_{22}, y_{22}), ..., (x_{2K}, y_{2K}) \}$ be a set of points in an image that we think roughly corresponds to those in the template. Note that the number of points in the two sets do not need to be the same. 
Point registration is the problem of aligning the points in $P_1$ (the template) over the points in $P_2$ (the image) \citep{besl1992method}. This is illustrated in Figure~\ref{fig:pointsetalignment} where we align template points $P_1$ (blue dots) with the image landmarks $P_2$ (red dots). We can intuitively understand this as finding a transformation function $T$ that applies some transform to all points in $P_1$ such that $T(P_1)$ aligns with $P_2$ in some distance metric $d$.
The point set registration methods differ in their choice of distance $d$ and the constraint they put on the transformation $T$. 
The non-rigid version of the CPD method in \citet{myronenko2010point} assumes that the points in the image $P_2$ are generated by a Gaussian distribution ``around'' the template points $P_1$ and it puts only mild regularity conditions on the transform $T$. In particular, it assumes that points move freely but coherently. \citet{myronenko2010point} model the problem using a gaussian mixture model. 
We rely directly on their algorithm to align our template to the image by supplying our template points $P_1$ and image landmarks $P_2$. The algorithm applies the standard Expectation Maximisation (EM) method to solve the maximum likelihood problem.

It has been suggested that the CPD method can be improved by using a neural network to learn the transformation \citep{li2019cpdnet} but we do not consider this here. There are also examples of more general learning based table segmentation that does not rely on a pre-specified template, see e.g. \citet{clinchant2018tables}.

\begin{figure}\centering
\includegraphics[width=0.7\textwidth]{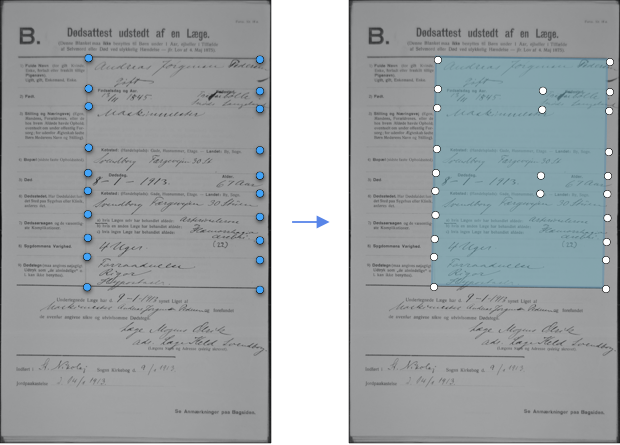}
\caption{An example of a set of points that define a template.}
\label{fig:templatepoints}
\end{figure}

\subsubsection{Results}
\begin{figure}\centering
\includegraphics[width=0.8\textwidth]{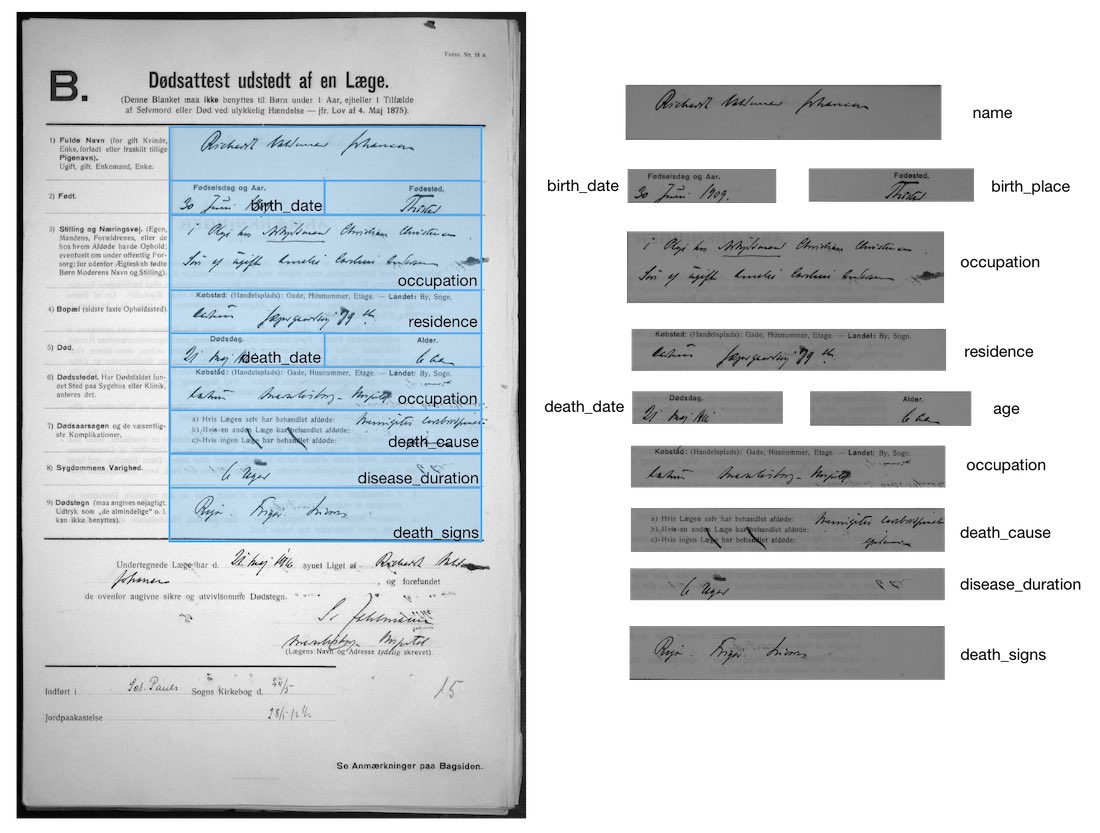}
\caption{Example of field images extracted using the fitted template.}
\label{fig:templatecuts}
\end{figure}

The initial template is constructed by finding a well-scanned death certificate where we can manually extract a number of points which can be used as anchors for the template, see Figure~\ref{fig:templatepoints}. This is only done once and these points comprise the template point set $P_1$.
After the template has been established, each type B death certificate is subjected to a set of morphological operations (erosion and dilation) to find pixels belonging to table rows or columns. The image is first thresholded (i.e. converted from gray scale into black/white where all pixels are either $0$ or $255$) and afterwards erosion and dilation operators are applied. Erosion and dilation are common non-probabilistic ways to extract straight (vertical or horizontal) lines in computer vision problems, see e.g. \citet{szeliski2010computer}. The morphological operations require careful hand-tuning to find the optimal parameters for the document type but these parameters only need to be tuned once for the whole collection. When the image has been reduced to straight white lines on a black background, we detect the intersection points of the lines using the Harris corner detection algorithm \citep{harris1988combined}. The detected corners provide a number of landmark points that inform us on the location, scale, rotation etc. of the table. The first two images in Figure~\ref{fig:dclandmarks} display a death certificate before and after application of the morphological operations, the third image shows the corners detected by the Harris algorithm, and the fourth image shows the final identified landmark points. 
The coordinates of the landmarks comprise the point set $P_2$. 
Next, we apply the CPD algorithm to align the template points in $P_1$ to the image landmarks in $P_2$. This iterative fitting process is illustrated in Figure~\ref{fig:pointsetalignment}. Even if extra landmark points are detected, the CPD algorithm remains robust as long as the detected landmark points $P_2$ are sufficiently spread out along the axes, i.e. if the noise is fairly uniform.
Once the points have been aligned, we use the estimated transformation $T$ to fit the template onto the image. The template then guides the cuts that separate the larger image into individual field images. This is illustrated in Figure~\ref{fig:templatecuts}. Figures~\ref{fig:incorrectpreddates}-\ref{fig:correctpreddates} show examples of the final segmented fields for dates. We do not report quantitative measures of segmentation performance but note that the end-to-end transcription accuracy in Section~\ref{sec:dc:crowd} will include any errors produced by the segmentation step if these make the field images unreadable.

The field images could be further segmented to limit noise and assure that the position and scale of the text is similar between samples. A possible way to implement this is a Mask R-CNN network \citep{he2017maskrcnn} which could be trained to predict regions of noise and handwritten text respectively. The text regions are then extracted and re-aligned on a blank image. 
However, this adds further complexity to the problem by introducing an additional segmentation model. The Mask R-CNN would need training data which requires manual annotation of text outlines to create a ground truth dataset. We will instead run the transcription model directly on the field images without any additional segmentation/cleaning and show that this is a viable approach.

\subsection{Step 3: Transcription}\label{sec:dc:transcription}
\subsubsection{Method}
Transcription is the process of converting an image of text into a string representation. Assume we have sequences of the form
$Y_i = (Y_{i,1}, Y_{i,2}, ..., Y_{i,T})$ where $T$ is the maximum sequence length and each $Y_{it}$ is a random variable on the sample space
$$\Omega = \{ \texttt{<Token1>}, \texttt{<Token2>}, ..., \texttt{<TokenN>} \}.$$
We call $\Omega$ a dictionary (or token space) and the tokens $\{ \texttt{<Token1>}, ..., \texttt{<TokenN>} \}$ serve as placeholders that can contain any information, be it individual characters, words, numbers, special symbols, or any combination thereof.\footnote{There are also sequence models relying on vector-space embeddings of words, e.g. the image captioning model of \citet{xu2015show}. The output from the learned mapping will then be a vector in $\mathbb{R}^K$ with $K$ being the embedding space dimension. This vector is then mapped to the closest (in some metric) known word in this space by a deterministic function, hence the final prediction will still be of the form $Y_i \in \Omega^T$. However, word embeddings are usually constructed by performing an auxiliary task on a large dataset of text, and it is not entirely obvious how such an approach would work for other things like cause-of-death or name, i.e. what properties would we expect from reasonable ``name'' embeddings? See also \citet{ye2019secret} who use twitter data to create an embedding space for names.} 
The contents of the dictionary depends on the application and there are essentially no restrictions.
For example, if $\Omega$ contains all characters in the alphabet then we can represent any conventional word of length $T$ as a sequence $Y_i  \in \Omega^T$. Similarly, if $\Omega$ contains whole words then we can represent a sentence of $T$ words equivalently as $Y_i  \in \Omega^T$. Based on an image $\textbf{X}_i$ we want to predict a sequence $Y_i = f(\textbf{X}_i)$ where $f$ is some unknown map from images to sequences. Learning the sequence mapping $f$ is the primary transcription problem. 

The optical character recognition (OCR), handwritten text recognition (HTR), and scene-text detection literature are all concerned with this problem, but the solutions differ heavily in how they model $f$ and how they choose their tokens, i.e. how they choose $\Omega$. Some methods rely on character-by-character transcription where $\Omega$ is simply the alphabet \citep{graves2008novel} and hence these models are, in theory, not limited in the set of words they can transcribe. Other approaches, like vector-space models, rely on (embeddings of) whole words such as the image captioning model in \citet{xu2015show}. In some cases, these choices limit the generality of the method, e.g. the whole-word models will not be able to transcribe words that are not in the dictionary. In our case, we are mainly interested in (1) numerical information such as dates, numbers, and area codes, and (2) string information such as names, locations and, cause-of-death. All of these fields have dictionaries that we can easily construct.
In the contemporary literature, it is common to use neural networks to model $f$ with various degrees of sophistication, see e.g. the introduction in \citet{graves2008novel}. The majority of the work has been on supervised methods where models are trained on pairs $(\textbf{X}_i, Y_i)$ but there are examples of unsupervised approaches as well, e.g. \citet{gupta2018learning}. 

We limit our attention to supervised models and focus on situations where the dictionary is small, well-defined and consists of a few  words and characters, i.e. a \emph{constrained} transcription model. Our intuition is that there is no reason to allow predictions such as "Bob" and "Apples" if we know that these should never occur in the data. For example, to represent any standard date we only need the digits $0-9$, names of the months, and some special characters.\footnote{This is less straightforward for transcription of names. However, it is possible to construct dictionaries based on the most common names and only transcribe these. Our initial tests of this approach have been promising for reading names in heavily cluttered and poorly segmented fields, but we do not present the results here.} 

\textbf{Model.} %
We utilise an attention-based neural network suggested by \citet{xu2015show} for image captioning and repurpose it for transcription of handwritten text. Originally, the model by \citet{xu2015show} predicts a string of words that describes the contents of an input image, e.g. if it is supplied with an image of a bird sitting in a tree then the model would predict the sentence ``A bird sitting in a tree''. However, the model of \citet{xu2015show} is generally applicable to tasks that involve image inputs and sequence outputs -- which is exactly what characterises the transcription problem. A similar attention-based model for OCR was proposed by \citet{lee2016recursive} and it has been demonstrated that more complex multi-head attention models can transcribe whole paragraphs of handwritten text (multiple lines) without prior line segmentation, see \citet{bluche2017scan}. 

In our implementation of \citet{xu2015show}, we replace the final embedding output with a softmax layer. This allows the model to predict probabilities for each of the tokens in our dictionary $\Omega$. In particular, assume that we have an image $\textbf{X}_i$ and we are at step $t$ in the prediction of the corresponding sequence $Y_i = (Y_{i,1}, ..., Y_{i,T})$. The model would predict the step $t$ conditional probabilities $$\pr{Y_{i,t} = k \, | \, \textbf{X}_i, Y_{i,t-1} = \hat{y}_{i,t-1}, ..., Y_{i,1} = \hat{y}_{i,1}}, \quad k \in \Omega$$ where $(\hat{y}_{i,t-1}, ..., \hat{y}_{i,1})$ are the previously predicted tokens in the sequence. 


\begin{figure}\centering
\includegraphics[width=0.45\textwidth]{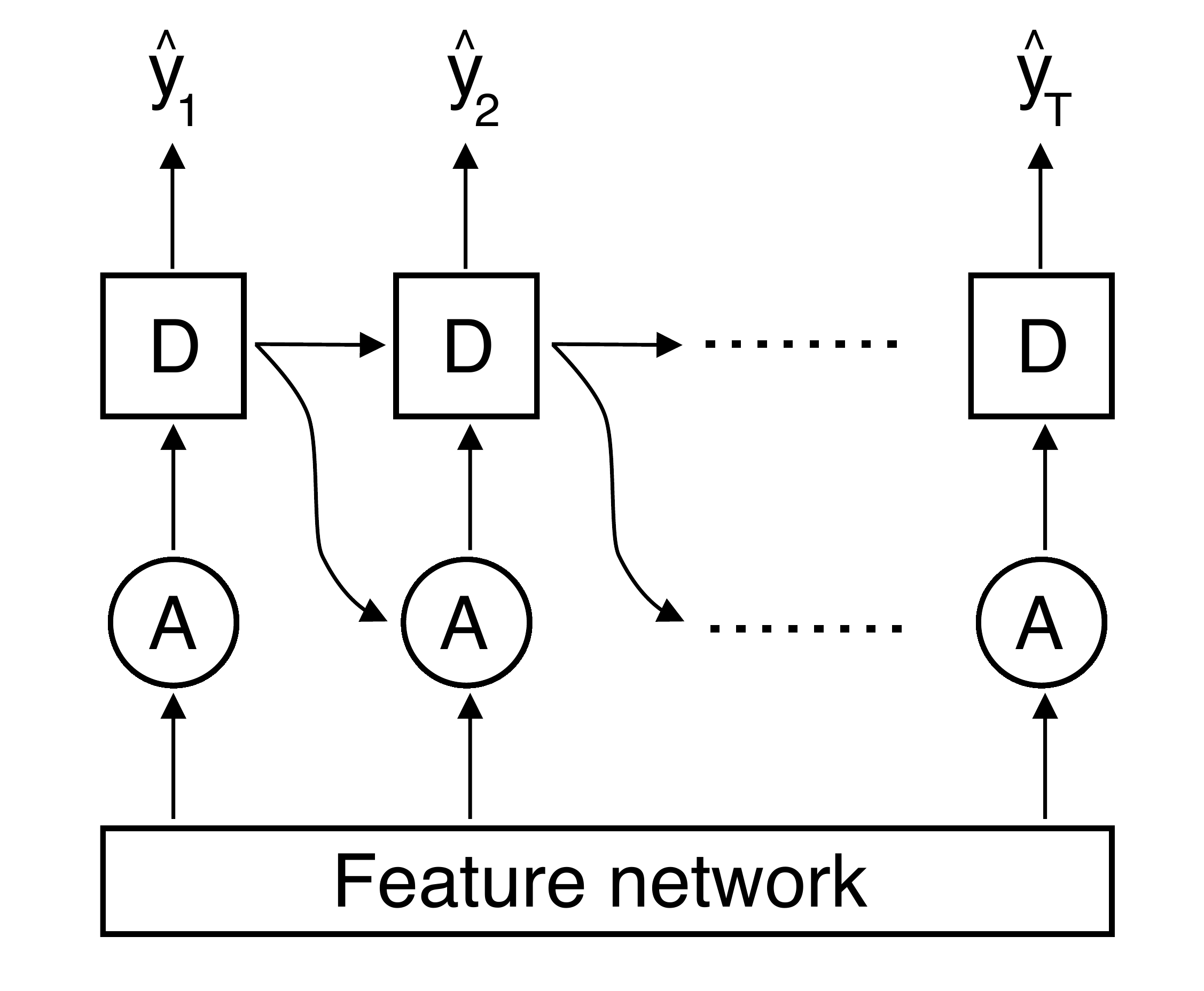}
\caption{Transcription model based on the image captioning model in \citet{xu2015show}. The feature network extracts features from the input image. The attention mechanism, denoted by $A$, weighs the features and supply these as input to an RNN cell, $D$, that predicts one token of the sequence. The hidden state of the RNN cell is updated, and the updated state is passed to the attention network to decide on the next weights for the feature map. The process repeats until an end-of-sequence prediction is made.}
\label{fig:transcriptionmodel}
\end{figure}

The model architecture is depicted in Figure~\ref{fig:transcriptionmodel}. First, the feature network encodes the input image into a feature map. The attention mechanism $A$ then applies soft-attention to this feature map by predicting weights in $[0, 1]$ and taking the element-wise product of the features and weights. As noted by \citet{xu2015show}, soft-attention is fully differentiable so we can use standard back-propagation to train the network.\footnote{In our work, we have found that the soft attention mechanisms are significantly easier to train compared to their hard attention counterparts. This is also noted by \citet{lee2016recursive}.} The attention-weighted feature map is passed to a recurrent neural network (RNN) cell $D$. This cell predicts the next token in the sequence and updates its hidden state. The hidden state is fed to the attention mechanism which updates the attention weights, i.e. ``where should we look next''? This supplies the recurrent cell with a new weighting of the feature map. The RNN cell uses the hidden state and the weighted features to predict the next token in the sequence. This procedure is repeated until an end-of-sequence token is predicted by the model.




An alternative approach is a neural network with the Connectionist Temporal Classification (CTC) loss. Such networks have achieved state-of-the-art results on unconstrained transcription \citep{graves2008novel}, i.e. in settings where there are no restrictions on the words the model can transcribe -- implying that $\Omega$ contains the alphabet. \citet{graves2006connectionist} suggested the CTC loss for sequence labelling. The CTC-based networks slice the entire input image into a sequence of feature vectors and for each feature vector make a prediction of the character/token that this particular slice contains. These predictions are then combined with the CTC loss function to handle duplicate predictions and ensure alignment of the predicted sequence with the ground truth during training. 
Usually, the CTC networks require more careful pre-processing and line segmentation. For example, \citet{graves2008novel} employ substantial line segmentation and cleaning before running the images through the transcription model. It is doubtful that these models would work well on the raw field images from the death certificates without further processing. We want to limit the amount of pre-processing as much as possible to simplify the pipeline and hence have not considered the CTC approach here. However, in ongoing work we are looking to apply CTC for token-based transcription and compare the attention-based model against the CTC-based US census transcription model being developed at the BYU Record Linking Lab.\footnote{See \url{https://rll.byu.edu/}} 

\textbf{Prediction.} Given an image $\textbf{X}_i$ we want to predict the corresponding sequence $Y_i \in \Omega^T$. At each sequence step $t$, we use the model to predict probabilities over the tokens in the dictionary $\Omega$. Due to dependence in the sequence, the predicted probabilities across the dictionary at step $t$ depend on all the previously predicted tokens at steps $(t-1, t-2, ..., 1)$. Exhaustive search over all possible combinations of tokens quickly becomes infeasible. As an example, the dictionary of the date transcription model contains 25 possible tokens, and with 11 elements in the sequence, we would have to enumerate and score $25^{11}$ different sequences for each sample we want to predict. Hence it is necessary to rely on greedy algorithms to find a feasible (but possibly suboptimal) solution. We use beam search as suggested by \citet{xu2015show}.

%


\textbf{Evaluation}. We evaluate the performance of the transcription models using two metrics. One relates to the average accuracy across individual tokens and the other to the accuracy of complete sequences. To fix notation let $y_i = (y_{i,1}, ..., y_{i,k_i})$ be a realised ground truth sequence and $\hat{y}_i = (\hat{y}_{i,1}, ..., \hat{y}_{i,h_i})$ a predicted sequence for image $\textbf{X}_i = \mathbf{x}_i, i = 1, ..., n$ where $n$ is the size of the evaluation dataset. Assume that the ground truth sequence is always padded so $k_i \geq h_i$ and that the padding value is chosen such that $y_{i,j} \not= \hat{y}_{i,j}$ for $j > k_i$. Moreover, let $I(\cdot)$ be the indicator function with some predicate. We define the Token Accuracy ($\text{TA}$) and $m$-error Sequence Accuracy ($\text{SA}_m$) by
\begin{align}
\text{TA} &= \frac{1}{\sum_{i} k_{i}} \sum_{i=1}^n \sum_{j=1}^{k_i} I(x_{i,j} = \hat{x}_{i,j}), \label{eq:metricta}\\[1em]
\text{SA}_m &= \frac{1}{n} \sum_{i=1}^n I\left( \sum_{j=1}^{k_i} I(x_{i,j} \not= x_{i,j}) \; \leq \; m \right). \label{eq:metricsa}
\end{align}
$\text{TA}$ has an interpretation as the average accuracy across all predicted tokens. $\text{SA}_m$ is the proportion of correctly predicted sequences when $m$ mistakes are allowed in the sequence. The token and sequence accuracies are closely related to the character and word accuracies in the HTR literature, see e.g. \citet{graves2008novel}. However, our dictionary contains tokens which can be chosen arbitrarily, i.e. a token can be a character or a word.

Note that both TA and SA are sensitive to alignment, so while a single substitution produces one sequence error, missing or adding an extra token would misalign the entire sentence. For example, in the following prediction the model misses a single token but this causes four errors in the predicted sequence because of misalignment:
\begin{center}\vspace{1em}
\begin{tabular}{rccccccc}
\textbf{Ground truth:} & \texttt{<Start>} & \texttt{<1>} & \texttt{<2>} & \texttt{<7>} & \texttt{<0>} & \texttt{<End>} \\
\textbf{Prediction:} & \texttt{<Start>} & \texttt{<1>} & \texttt{<7>} & \texttt{<0>} & \texttt{<End>} & \texttt{<Pad>} \\
& Correct & Correct  & Error & Error & Error & Error
\end{tabular}\vspace{1em}
\end{center}
Hence missing a token can impose a heavy accuracy penalty when using these measures.  Another possibility for measuring performance would be various string distances, e.g. the Levenshtein distance. These can be adapted to tokens instead of characters and do not have the alignment problem. We do not consider this futher but note that relying on string distance measures would only improve the metrics for our models.  

\subsubsection{Setup and training}
We train separate models for dates and ages but both are based on the same attention neural network. Below we detail the choices of dictionary, hyperparameters and the training procedure used to train the two models on the type B death certificates.
 
\textbf{Dictionary.} The age transcription model uses a dictionary consisting of the digits $(0, 1, ..., 9)$ together with start and end markers to delimit the sequence. The final age dictionary is $\Omega_{age} = \{ \texttt{<Start>}, \texttt{<0>}, \texttt{<1>}, ..., \texttt{<9>}, \texttt{<End>} \}$ which can represent integer ages of arbitrary length, in principle allowing for 3-digit ages. For example, $12$ would be tokenised as
\begin{mdframed}[nobreak=false,align=center]
\texttt{\texttt{<Start>}, \texttt{<1>}, \texttt{<2>}, \texttt{<End>}}
\end{mdframed}
The dictionary excludes ratios of years, e.g. $1/2$, and any suffixes such as years, months, days, and hours. It would be straightforward to add additional tokens but since our ground truth transcriptions do not contain this information we cannot train the model to recognise them.

The date transcription model uses a larger dictionary. Each month is considered a separate token, \texttt{<January>, <February>, ..., <December>}, and all digits are separate tokens, \texttt{<0>, <1>, ..., <9>}. We include two separator tokens, \texttt{<DayMonthSeparator>} and \texttt{<MonthYearSeparator>}, which encode the separators between the day--month and month--year components of the dates. 
We also include sequence delimiters \texttt{<Start>, <End>} for marking the start/end of the sequence and padding \texttt{<Pad>}. The entire dictionary $\Omega_{date}$ is given in Table~\ref{table:tokenspacedates} in the Appendix and it provides a standardised representation of any date. For example, the date \texttt{1/7-2010} would be tokenised as
\begin{mdframed}[nobreak=false,align=center] \texttt{<Start>,<1>,<DayMonthSeparator>,<July>,<MonthYearSeparator>,<2>,<0>,<1>,} \\ \texttt{<0>,<End>}
\end{mdframed}
Similarly, the date \texttt{September 20th, 90} would by tokenised as
\begin{mdframed}[nobreak=false,align=center]\texttt{<Start>,<2>,<0>,<DayMonthSeparator>,<September>,<MonthYearSeparator>,} \\ \texttt{<9>,<0>,<End>}
\end{mdframed}

\textbf{Data.} The date model is trained on 11,630 manually transcribed dates. The age model is trained on 11,072 manually transcribed ages. To boost the amount of training data, we apply an augmentation procedure to create multiple distorted versions of each original training sample. 
Prior to training, the mean and variance are estimated using a sample of $20,000$ augmented images. The mean and variance are then used to standardise the pixel values in each training image. The ground truth sequences (respectively for dates and ages) are tokenised according to the defined dictionaries and they are padded to all have same length $T$. For age this length is $T = 4$ tokens while it is $T = 11$ tokens for dates. During training the models are fed batches of pairs $\{ (\textbf{x}_i, y_i) \}_i$ where each training pair consists of a field image $\mathbf{x}_i$ and its corresponding ground truth sequence $y_i$. 

\textbf{Hyperparameters.} The transcription models require various hyperparameters that must be selected by the researcher. Some parameters are intuitive, e.g. the size of the input images, but others are more abstract such as the number of hidden nodes in a specific layer of the neural network. There is limited guidance for choosing these parameters. For the best performance, it is necessary to do hyperparameter optimisation where several sets of parameters are evaluated on a tuning dataset. We do not consider this and instead use manually selected values that yield acceptable results on the death certificates.

The feature network is the convolutional part of a pre-trained ResNet-101 network \citep{he2016deep} which we tune during training with a learning rate of $0.0003$. The encoded image size is $(6,20)$ which is obtained by applying adaptive max pooling after the final layer of the ResNet-101 feature network. The learning rate is $0.0005$ for the remaining parameters (i.e. all parameters not in the feature network). The learning rate is lower for the feature network to combat over-fitting as the ResNet-101 network is deep. In addition, we use learning rate decay with a factor of $0.25$ which is applied to both the main and feature network learning rates at steps $10,000$, $15,000$ and $25,000$.

The RNN cell is a Gated Recurrent Unit (GRU) \citep{cho2014gru}. The number of hidden units in the attention network is $512$ while it is $2,056$ in the RNN cell. The initial hidden state of the RNN cell is learned from the features using a fully-connected layer.
We apply dropout with probability $0.5$ before the final softmax prediction layer. The model is trained with back-propagation using the Adam optimiser \citep{kingma2014adam}. The batch size is 42 images which is the maximum possible on our hardware. 

The date model is trained for 45,000 steps until stagnating improvement in the loss function. This is equivalent to the model encountering each training sample on average 248 times (also known as 248 epochs). The age model is trained for $28,000$ steps implying that each sample is encountered on average 162 times. Note that the training datasets are augmented; while the model does encounter the same original sample multiple times, the sample will have different random distortions.



\subsubsection{Results}
We evaluate the performance of the transcription models using the $\text{TA}$ and $\text{SA}_m$ metrics from Equations~\ref{eq:metricta}--\ref{eq:metricsa}. This section focuses on the performance of the ML transcription models in isolation, so all results are conditional on the transcription models receiving perfectly segmented images.
The trained transcription models are used to predict the images in the date and age evaluation datasets using beam search. Prior to prediction, the images are standardised to zero mean and unit variance using the mean and variance estimated from the entire evaluation dataset.\footnote{This is common practice and as straightforward as it sounds. We take the average and standard deviation over all pixels in all images, and then respectively subtract and divide each pixel by these values.} No other pre-processing is applied. Transcription of a single image happens in less than 75ms. 

\begin{table}\centering
\begin{tabular}{l rrrr @{\extracolsep{8pt}} r @{} r @{} r @{}}
\toprule\toprule
& \multicolumn{4}{c}{Date} & \multicolumn{3}{c}{Age} \\ \cline{2-5}\cline{6-8}
 & $TA$ & $SA_0$ & $SA_1$ & $SA_2$  & $TA$ & $SA_0$ & $SA_1$ \\ \midrule


Real 					& .922 & .661 & .933 & .981 & .966 & .936 & .988 \\
					& $(.008)$ & $(.015)$ & $(.008)$ & $(.004)$ & $(.006)$ & $(.008)$ & $(.003)$ \\[0.25em]

Real w. augmentation 	& .979 & .905 & .989 & .991 & .985 & .972 & .999  \\
					& $(.005)$ & $(.009)$ & $(.003)$ & $(.003)$ & $(.004)$ & $(.005)$ & $(.001)$  \\[0.25em]

\bottomrule\bottomrule
\end{tabular}
\captionof{table}{Token (TA) and Sequence (SA) Accuracy on ground truth evaluation set. 
Standard errors are given below each accuracy rate.
The first row uses only ground truth training samples, while the second row shows the result when training is conducted on the augmented dataset. The date training set contains $11,320$ samples, the evaluation set contains $1,000$ samples. The age training set contains $11,072$ samples, the evaluation set contains $1,000$ samples. The dates contain both birth and death dates. Note that these are not end-to-end accuracy rates, so they do not factor in the performance of the table segmentation (i.e. segmentations that obscures the written information have been discarded). The accuracy excludes the \texttt{<Start>} token but includes the \texttt{<End>} token (the \texttt{<Start>} token is forced in the network, so it will always be present).} 
\label{table:dc_transcription_eval}
\end{table}

\begin{figure}\centering
\includegraphics[width=0.9\textwidth]{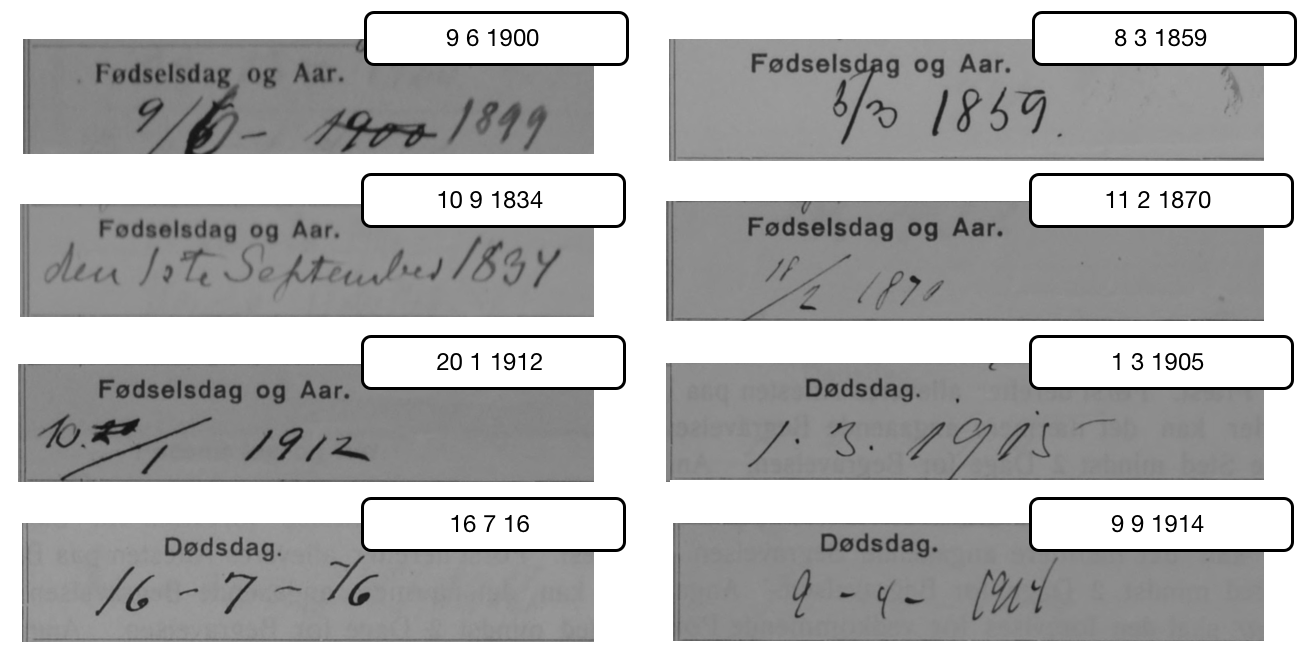}
\captionof{figure}{Random sample of incorrect predictions from the date transcription model. The label in the upper right corner displays the model prediction in the format day--month--year.}\vspace{3em}
\label{fig:incorrectpreddates}
\includegraphics[width=0.9\textwidth]{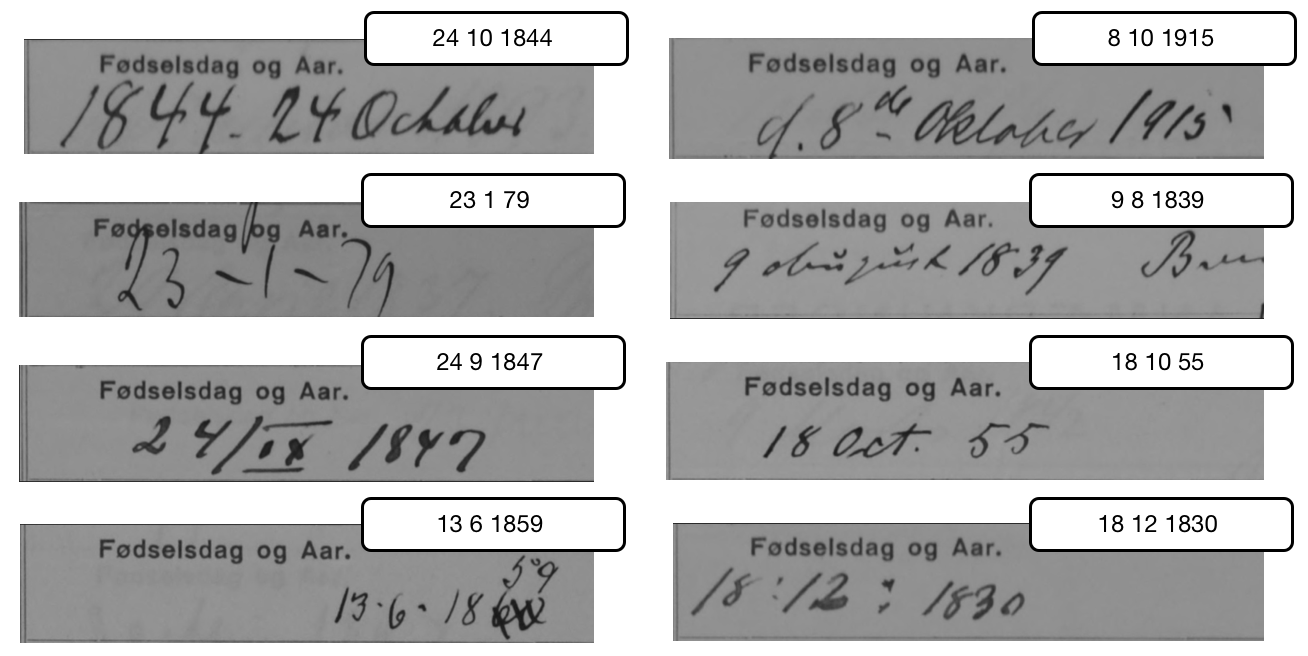}
\captionof{figure}{Random sample of correct predictions from the date transcription model. The label in the upper right corner displays the model prediction in the format day--month--year.}
\label{fig:correctpreddates}
\end{figure}

Table~\ref{table:dc_transcription_eval} shows the performance of the two transcription models -- for date and age respectively -- on their corresponding evaluation sets. We present results both for training with and without augmentation of the training dataset. The token accuracy (TA) for dates is $97.9 \%$ with augmentation and $92.2 \%$ without. For ages the TA is $98.5 \%$ with augmentation and $96.6 \%$ without. These are the average accuracies of predicting a single token correctly. Figures~\ref{fig:incorrectpreddates}-\ref{fig:correctpreddates} show random samples of incorrect and correct dates as predicted by the ML date transcription model.

In Table~\ref{table:dc_transcription_eval}, the zero-error sequence accuracies $SA_0$ are significantly lower than the token accuracies for both models. Under augmentation, the date model achieves an $SA_0$ of $90.5 \%$ while the age model has an $SA_0$ of $97.2 \%$. In the context of US censuses, \citet{nion2013census} transcribed age at a sequence accuracy of approximately $85 \%$ using convolutional neural networks. In ongoing (unpublished) work, the BYU Record Linking Lab is using a CTC-based approach to correct mistakes in US censuses, but we do not yet have performance metrics to compare against.

Notice that a sequence prediction is only correct if all tokens are predicted correctly. This implies correctness of $11$ tokens for dates and $4$ tokens for ages (including \texttt{<Start>} and \texttt{<End>} markers). 
Hence it is no surprise that the zero-error sequence accuracy is much higher for age than date. Also, the variation in dates is larger compared to ages with respect to both format and combination of digits/characters. 

It is apparent from Table~\ref{table:dc_transcription_eval} that an increase in the number of allowed mistakes per sequence improves the accuracy significantly. E.g. if we allow one mistake (i.e. one substitution) in the date sequence, the one-error sequence accuracy is $98.9 \%$. Under some circumstances, e.g. linking, it might be acceptable with a certain number of mistakes in the sequence. Also, in statistical models, the transcription errors might not matter unless they depend systematically on the transcribed information.


Notice also the difference between using augmentation and not in the first and second row of Table~\ref{table:dc_transcription_eval}. The significant differences are due to the small training datasets of $11,630$ dates and $11,072$ ages. If our training datasets were larger, the payoff from augmentation would be smaller. However, the differences display the benefits of augmentation to boost performance in smaller training datasets. 


\subsection{Evaluation of crowdsourcing and complete ML pipeline}\label{sec:dc:crowd}
We consider a comparison to a crowdsourced dataset. The usefulness of this comparison is twofold: (1) it allows us to evaluate if the ML-approach produces transcriptions that are on-par with those obtainable from crowdsourcing, and (2) it provides a measure of the end-to-end performance of the ML transcription pipeline, including errors of segmentation, transcription, and to an extent layout classification.\footnote{It does not provide us with a good measure of the recall of the layout classifications as we only consider certificates that are in both datasets. There might be type B certificates that (1) have not been detected by our ML model and (2) are not in the crowdsourced dataset. These will be missed here.}

\begin{table}\centering
\begin{tabular}{l cc @{\extracolsep{8pt}} cc cc cc}
\toprule\toprule
& \multicolumn{2}{c}{Date} & \multicolumn{2}{c}{Day} & \multicolumn{2}{c}{Month} & \multicolumn{2}{c}{Year} \\ \cline{2-3}\cline{4-5}\cline{6-7}\cline{8-9}
 & \colmarker{ML} & \colmarker{Crowd} & \colmarker{ML} & \colmarker{Crowd} & \colmarker{ML} & \colmarker{Crowd} & \colmarker{ML} & \colmarker{Crowd} \\
  & \colmarker{[1]} & \colmarker{[2]} & \colmarker{[3]} & \colmarker{[4]} & \colmarker{[5]} & \colmarker{[6]} & \colmarker{[7]} & \colmarker{[8]} \\ \midrule
$\text{SA}_0$ 			& .905 		& .963 		& .960 		& .983 & .970 & .987 & .972 & .988  \\
 				& $(.009)$ 	& $(.004)$ 	& $(.006)$ 	& $(.002)$ & $(.005)$ & $(.002)$ & $(.005)$ & $(.002)$  \\
\bottomrule\bottomrule
\end{tabular}
\captionof{table}{Zero-error Sequence Accuracy $\text{SA}_0$ for the ML data model (trained with augmentation) and crowdsourcing both evaluated on our date ground truth datasets. Second row contains the standard error of the accuracy rate. The evaluation set for the ML model consists of 1,000 samples, while the evaluation set for the crowdsourced predictions consists of 2,864 samples as we can pool both our training and evaluation ground truth datasets in this case. The training and evaluation sets have been twice manually reviewed and dates that are unreadable due to bad segmentation have been removed. Age is not directly transcribed in the crowdsourced dataset and hence excluded here. Column 1-2 is sequence accuracy for the whole date, while column 3-8 are sequence accuracy on the individual components of the date (day, month and year). Note that the comparison takes into account common date formatting, e.g. that the dates 01-10-2000 and 1-10-2000 convey the same point in time.} 
\label{table:csmleval}
\end{table}

Our own evaluation dataset -- as used in Table~\ref{table:dc_transcription_eval} -- was manually reviewed to exclude (1) images where segmentation errors obscured the text in the image, and (2) where the image belonged to a document of the wrong layout type (i.e. anything other than type B death certificates). Evaluating the model on this dataset provides a clean measure of the transcription model in isolation. However, it does give any insights on the transcription performance when the model might receive a badly segmented image. The crowdsourced dataset is different in this respect as it is constructed by humans looking at the raw document, finding the date field, and transcribing the text. This implies that the crowdsourced transcriptions cannot be impacted by segmentation or layout classification errors. By running the entire pipeline on the raw images and comparing the final transcription output to the crowdsourced transcription, we can get a measure of the overall performance of the pipeline in practice. Of course, this relies on the assumption that the crowdsourced data are perfectly transcribed which is not likely to be the case.

To get a baseline indication of the quality of the crowdsourced dataset, we compare it against our ground truth training and evaluation sets (overlap of 2,864 documents) and find that the dates are identical in 96.30\% of the cases, see Table~\ref{table:csmleval}. For comparison, the performance of the transcription model on the evaluation set (i.e. perfectly segmented images) is $90.5\%$. If we look at the individual components of the date, the ML performance is 96\%, 97\% and 97.2\% on days, months and years respectively. For crowdsourcing, the corresponding accuracy rates are 98.3\%, 98.7\% and 98.8\%. Thus, on the individual date components, the accuracies of crowdsourcing and ML are fairly close (although statistically different). However, this neglects the impact of the layout classification and segmentation steps in the pipeline. 


Next, we evaluate the ML transcriptions by using the crowdsourced transcriptions as ground truth. The crowdsourced transcriptions overlap with our sample for $23,263$ documents -- each containing a birth and death date for a total of $46,526$ dates -- and we filter out any overlap with the training sample used to train the ML model. Note that the crowdsourced dataset does not contain transcriptions of age. The dates predicted by the ML model and crowdsourcing are identical in 83.66\% of the cases\footnote{Not completely identical, we take common differences in formatting into account, so e.g. 01-3-2000 and 1/3-2000 would be considered equal.} and for 89.96\% of the dates the difference is less than one calendar year. This is a substantial difference compared to Table~\ref{table:dc_transcription_eval} where the ML sequence accuracy was 90.5\%. 
As elaborated above, the performance difference relative to Table~\ref{table:dc_transcription_eval} stems from two sources: (1) noise in the crowdsourced dataset, and (2) the other pipeline steps prior to transcription. Thus, unless (1) is large, this gives an approximation to the end-to-end performance of the whole pipeline. We should keep in mind that the 83.66\% sequence accuracy allows for zero mistakes in the predicted sequence and that this is the expected performance if we -- without any pre-processing or adjustments -- feed a collection of raw scans to the pipeline. Also, given the noise in the crowdsourced dataset, we can argue that 83.66\% might be a slightly conservative estimate unless crowdsourcing participants and ML make the exact same mistakes on the exact same documents.

Using the ML approach, it is cheap to transcribe additional fields on the documents as it only requires a training sample. As we have seen, the training sample can be much smaller than the full collection of documents. This can be exploited to produce higher sequence accuracy rates if there are internal correspondences between the fields in the source document. For example, the death certificates contain both birth and death date and age. These three fields should be internally consistent. If they do not match then either (1) the source document contains a mistake or (2) the ML model made a mistake. If we transcribe both age and dates and exploit the correspondence between these fields, we can filter out 5,767 cases where the predicted and implied age differ by more than one year. This leaves 17,496 documents where we achieve an ML sequence accuracy of 93.56\% end-to-end. Even if the problematic documents need to be manually transcribed, the ML model still produces a reduction in the manual transcription burden by around 75\% relative to manually transcribing the whole dataset of $23,263$ documents ($46,526$ field images). Hence, relationships between fields can provide automatic verification and be used to flag problematic records for manual review. 
This method can of course also be applied in a manual/crowdsourcing context, but in this case transcription of additional fields is more. 

\begin{figure}\centering
\hspace{-0.5em}\includegraphics[width=0.975\textwidth]{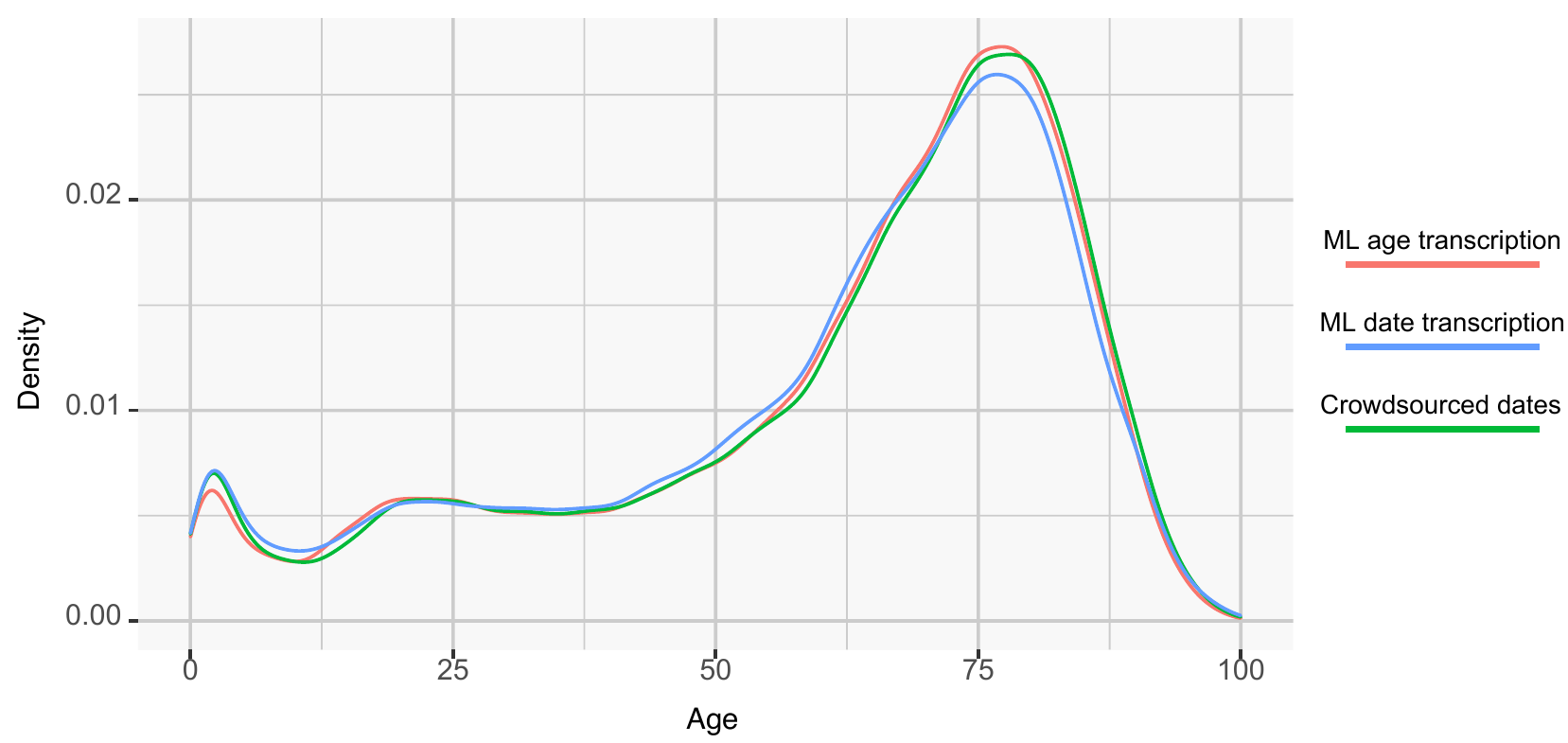}
\caption{Lifetime distributions. Kernel density estimates of the lifetime distribution implied by ML age transcriptions (red), ML date transcriptions (blue), and crowdsourced date transcriptions (green) in an overlapping sample of $23,263$ individuals restricted to the age interval $[0; 100]$. \emph{Implied age} refers to the difference between the birth and death dates in years.
}
\label{fig:kddist}
\end{figure}

In addition to the accuracy rates, we also compare the data obtained from the ML and crowdsourcing approaches. Any systematic bias in the ML model would produce deviations from the empirical distribution observed in the crowdsourced dataset. Also -- based on the discussion above -- we expect internal consistency between ML transcribed ages and dates. Figure~\ref{fig:kddist} compares kernel density estimates of the age distribution produced by (red line) ML age transcriptions, (blue line) ML date transcriptions, and (green line) crowdsourced date transcriptions. Note that the figure only displays ages in the interval $[0; 100]$ -- any ages outside this interval are discarded. In addition, we discard all predicted dates where the year does not contain four digits. For some dates in the source documents, the year has been abbreviated to only the last two digits (e.g. 1890 becomes 90 or 1910 becomes 10). We cannot easily infer the first two digits, as most of the death certificates span the time 1850--1950 so the leading digits could be either $18$ or $19$. This is not a shortcoming of the transcription model but rather a lack of information in the source documents.
In Figure~\ref{fig:kddist}, we see that the distribution of ML age transcriptions is very close to the age distribution implied by the crowdsourced date transcriptions. The age distribution implied by the ML date transcriptions also appear similar, although with a notable difference around ages 75--85. This deviation is not surprising as the dates contain longer sequences to transcribe (relative to age) and the ML model needs to transcribe both birth and death date correctly (at least down to the year) to get an approximately correct age prediction. 

Motivated by the notable difference in performance between the whole pipeline and transcription only, we manually review some of the cases where the crowdsourced and model predictions differ. This reveals, albeit qualitatively, that most discrepancies are related to segmentation issues where the CPD segmentation obscures the year in the date fields. The CPD segmentation template makes an arbitrary cut to separate the age and death date fields. The year is the rightmost component of the date and hence most likely to be impacted by this cut -- refer to the positioning of the death date and age fields in Figure~\ref{fig:typebfields}. This explanation is corroborated if we look at the accuracy rates for the individual date components with the crowdsourced dataset as ground truth: Day accuracy is 95.92\%, month accuracy is 96.98\%, and year accuracy is 89.11\%. Clearly, the accuracy for the year component is notably lower. When discovered, such issues can easily be corrected in the ML pipeline. Had the transcription been done manually, it would be much more costly to correct systematic transcription errors.

As a concluding remark, keeping in mind the resources and time needed to perform manual or crowdsourced transcription, we note that the ML end-to-end accuracy of 83.66\% -- trained on only 11,630 dates and 11,072 ages -- might in many cases be acceptable, especially for large document collections that are otherwise infeasible to transcribe. 

It should be noted that the ML model in this application has not been carefully optimised and it has only been trained on a small training dataset. It is conceivable that the model can perform substantially better. 
In addition, improvements in segmentation would also positively impact the end-to-end accuracy. The previously discussed within-field segmentation using Mask R-CNN \citep{he2017maskrcnn} would be an option for this. 
Also, in practical applications, the model can be used to speed up transcription while retaining manual review of each (or some) predictions. The reviewed predictions can then be used to re-train and improve the model. 

There is also the possibility to rely on the confidence predicted by the model and remove transcriptions where the model is highly uncertain. We could then apply a threshold on the prediction probabilities and manually review sequences with predicted probabilities below the threshold. Although, it is questionable if these probabilities can be understood as a general measure of confidence, see e.g. the discussion of calibration in \citet{guo2017calibration}. 

\section{Concluding remarks}\label{sec:conclusion}
We have provided two motivating examples that apply ML to collect data from scanned documents. First, we have shown that (unsupervised) layout classification can be a useful tool in intervention studies where treatment compliance is inferred from documents. As an exploratory step, the layout classification can reveal the composition of pages within the collection and details about assignment. We demonstrated on a clean dataset of nurse journals that a simple unsupervised method can achieve precision and recall around $1.0$ on an evaluation set of 4,000 journals with a minimum of manual annotation. This revealed non-compliance that would not have been discovered unless the whole collection (261,926 pages) was manually reviewed. 

Secondly, we have shown that an attention-based model is a viable method for transcribing a variety of handwritten information when we have rough segmentation at the field-level. The model works without the need for line detection or other time consuming pre-transcription steps. The end-to-end accuracy for transcribing birth and death dates was around 83\%. By combining transcription of multiple fields (age and dates) for verification, the end-to-end accuracy increased to approximately 93.56\%. However, this verification procedure left approximately 25\% of the documents for manual review. This performance is close to the 96.3\% sequence accuracy we observed for crowdsourced data, but the ML model still provided a significant reduction in the manual transcription burden. In isolation, the ML transcription step had an accuracy of around 96\% on a perfectly segmented evaluation set.
In closing, we emphasise that the ML end-to-end accuracy rates are scalable to collections of arbitrary size with limited resource consumption. The cost difference between transcribing $100,000$ and 2 million documents is negligible as opposed to the cost of the equivalent manual transcription.

\newpage 
\singlespacing

\nocite{*}
\printbibliography

\newpage 
\doublespacing

\section*{Appendix}
\subsection*{Dates dictionary}
\begin{table}[h!]
\begin{tabular}{rl}
\toprule\toprule
Description & Tokens \\ \midrule
Day and year digits & \texttt{<0>, <1>, <2>, <3>, <4>, <5>, <6>, <7>, <8>, <9>} \\
Months & \texttt{<January>, <February>, <March>, <April>} \\ & \texttt{<May>, <June>, <July>, <August>, <September>} \\ & \texttt{<October>, <November>, <December>} \\
Separators & \texttt{<DayMonthSeparator>, <MonthYearSeparator>} \\
Sequence markers and padding & \texttt{<Padding>, <Start>, <Stop>} \\
\bottomrule\bottomrule
\end{tabular}
\caption{Dictionary $\Omega_{date}$ for the date transcription model.}
\label{table:tokenspacedates}
\end{table}



\end{document}